\documentclass[final,5p,times,twocolumn,authoryear]{elsarticle}
\usepackage[hidelinks]{hyperref}
\usepackage{lineno}
\hypersetup{colorlinks=true, linkcolor=blue, filecolor=magenta, urlcolor=cyan}
\usepackage{amsmath}
\journal{Elsevier}

\begin{document}
\begin{frontmatter}
\title {Multi-branch Convolutional Neural Network for Multiple Sclerosis Lesion Segmentation}
\address[a]{Pattern Analysis and Computer Vision (PAVIS), Istituto Italiano di Tecnologia (IIT), Genoa, Italy}
\address[b]{Science and Technology for Electronic and Telecommunication Engineering, University of Genoa, Italy}
\address[c]{Human Neuroscience Platform, Fondation Campus Biotech Geneva, Switzerland}
\address[d]{Neuroimaging Research Unit, Institute of Experimental Neurology (INSPE), Division of Neuroscience, San Raffaele Scientific Institute, Milan, Italy}
\address[e]{Dipartimento di Informatica, University of Verona, Italy}
\address[f]{NeuroInformatics Laboratory, Fondazione Bruno Kessler, Trento, Italy}
\author[a,b]{Shahab Aslani\corref{corr1}}\ead{shahab.aslani@iit.it}
\cortext[corr1]{Corresponding authors}
\author[a,c]{Michael Dayan}
\author[d]{Loredana Storelli}
\author[d]{Massimo Filippi}
\author[a,e]{Vittorio Murino}
\author[d]{Maria A Rocca}
\author[a,f]{Diego Sona\corref{corr1}}\ead{diego.sona@iit.it}
\begin{abstract}
In this paper, we present an automated approach for segmenting multiple sclerosis (MS) lesions from multi-modal brain magnetic resonance images. Our method is based on a deep end-to-end 2D convolutional neural network (CNN) for slice-based segmentation of 3D volumetric data. The proposed CNN includes a multi-branch downsampling path, which enables the network to encode information from multiple modalities separately. Multi-scale feature fusion blocks are proposed to combine feature maps from different modalities at different stages of the network. Then, multi-scale feature upsampling blocks are introduced to upsize combined feature maps to leverage information from lesion shape and location. We trained and tested the proposed model using orthogonal plane orientations of each 3D modality to exploit the contextual information in all directions. The proposed pipeline is evaluated on two different datasets: a private dataset including 37 MS patients and a publicly available dataset known as the ISBI 2015 longitudinal MS lesion segmentation challenge dataset, consisting of 14 MS patients. Considering the ISBI challenge, at the time of submission, our method was amongst the top performing solutions. On the private dataset, using the same array of performance metrics as in the ISBI challenge, the proposed approach shows high improvements in MS lesion segmentation compared with other publicly available tools.
\end{abstract}


\end{frontmatter}

\section{Introduction}
\label{intro}
Multiple sclerosis (MS) is a chronic, autoimmune and demyelinating disease of the central nervous system causing lesions in the brain tissues, notably in white matter (WM) \citep{steinman1996multiple}. Nowadays, magnetic resonance imaging (MRI) scans are the most common solution to visualize these kind of abnormalities owing to their sensitivity to detect WM damage \citep{compston2008multiple}.

Precise segmentation of MS lesions is an important task for understanding and characterizing the progression of the disease \citep{rolak2003multiple}. To this aim, both manual and automated methods are used to compute the total number of lesions and total lesion volume. Although manual segmentation is considered the gold standard \citep{simon2006standardized}, this method is a challenging task as delineation of 3-dimensional (3D) information from MRI modalities is time-consuming, tedious and prone to intra- and inter-observer variability \citep{sweeney2013oasis}. This motivates machine learning (ML) experts to develop automated lesion segmentation techniques, which can be orders of magnitude faster and immune to expert bias.

Among automated methods, supervised ML algorithms can learn from previously labeled training data and provide high performance in MS lesion segmentation. More specifically, traditional supervised ML methods rely on hand-crafted or low-level features. For instance, \citet{cabezas2014boost} exploited a set of features, including intensity channels (fluid-attenuated inversion-recovery (FLAIR), proton density-weighted (PDw), T1-weighted (T1w), and T2-weighted (T2w)), probabilistic tissue atlases (WM, grey matter (GM), and cerebrospinal fluid (CSF)), a map of outliers with respect to these atlases \citep{schmidt2012automated}, and a set of low-level contextual features. A Gentleboost algorithm \citep{friedman2000additive} was then used with these features to segment multiple sclerosis lesions through a voxel by voxel classification.

During the last decade, deep learning methods, especially convolutional neural networks (CNNs) \citep{lecun1998gradient}, have demonstrated outstanding performance in biomedical image analysis. Unlike traditional supervised ML algorithms, these methods can learn by themselves how to design features directly from data during the training procedure \citep{lecun2015deep}. They provided state-of-the-art results in different problems such as segmentation of neuronal structures \citep{ronneberger2015u}, retinal blood vessel extraction \citep{liskowski2016segmenting}, cell classification \citep{han2016hep}, brain extraction \citep{kleesiek2016deep}, brain tumor \citep{havaei2017brain}, tissue \citep{moeskops2016automatic}, and MS lesion segmentation \citep{valverde2017improving}.

In particular, CNN-based biomedical image segmentation methods can be categorized into two different groups: patch-based and image-based methods. In patch-based methods, a moving window scans the image generating a local representation for each pixel/voxel. Then, a CNN is trained using all extracted patches, classifying the central pixel/voxel of each patch as a healthy or unhealthy region. These methods are frequently used in biomedical image analysis since they considerably increase the amount of training samples. However, they suffer of an increased training time due to repeated computations over the overlapping features of the sliding window. Moreover, they neglect the information over the global structure because of the small size of patches \citep{tseng2017joint}. 

On the contrary, image-based approaches process the entire image exploiting the global structure information \citep{tseng2017joint, brosch2016deep}. These methods can be further categorized into two groups according to the processing of the data: slice-based segmentation of 3D data \citep{tseng2017joint} and 3D-based segmentation \citep{brosch2016deep}. 

In slice-based segmentation methods, each 3D image is converted to its 2D slices, which are then processed individually. Subsequently, the segmented slices are concatenated together to reconstruct the 3D volume. However, in almost all proposed pipelines based on this approach, the segmentation is not accurate, most likely because the method ignores part of the contextual information \citep{tseng2017joint}.

In 3D-based segmentation, a CNN with 3D kernels is used for extracting meaningful information directly from the original 3D image. The main significant disadvantage of these methods is related to the training procedure, which usually fits a large number of parameters with a high risk of overfitting in the presence of small datasets. Unfortunately, this is a quite common situation in biomedical applications \citep{brosch2016deep}. To overcome this problem, recently, 3D cross-hair convolution has been proposed \citep{liu2017triple, tetteh2018deepvesselnet}, where three 2D filters are defined for each of the three orientations around a voxel (each one is a plane orthogonal to X, Y, or Z axis). Then, the sum of the result of the three convolutions is assigned to the central voxel. The most important advantage of the proposed idea is the reduced number of parameters, which makes training faster than a standard 3D convolution. However, compared to standard 2D convolution (slice-based), still, there are three times more parameters for each layer, which increases the chance of overfitting in small datasets.

\begin{figure}[t!]
\centering
\includegraphics[width=85mm]{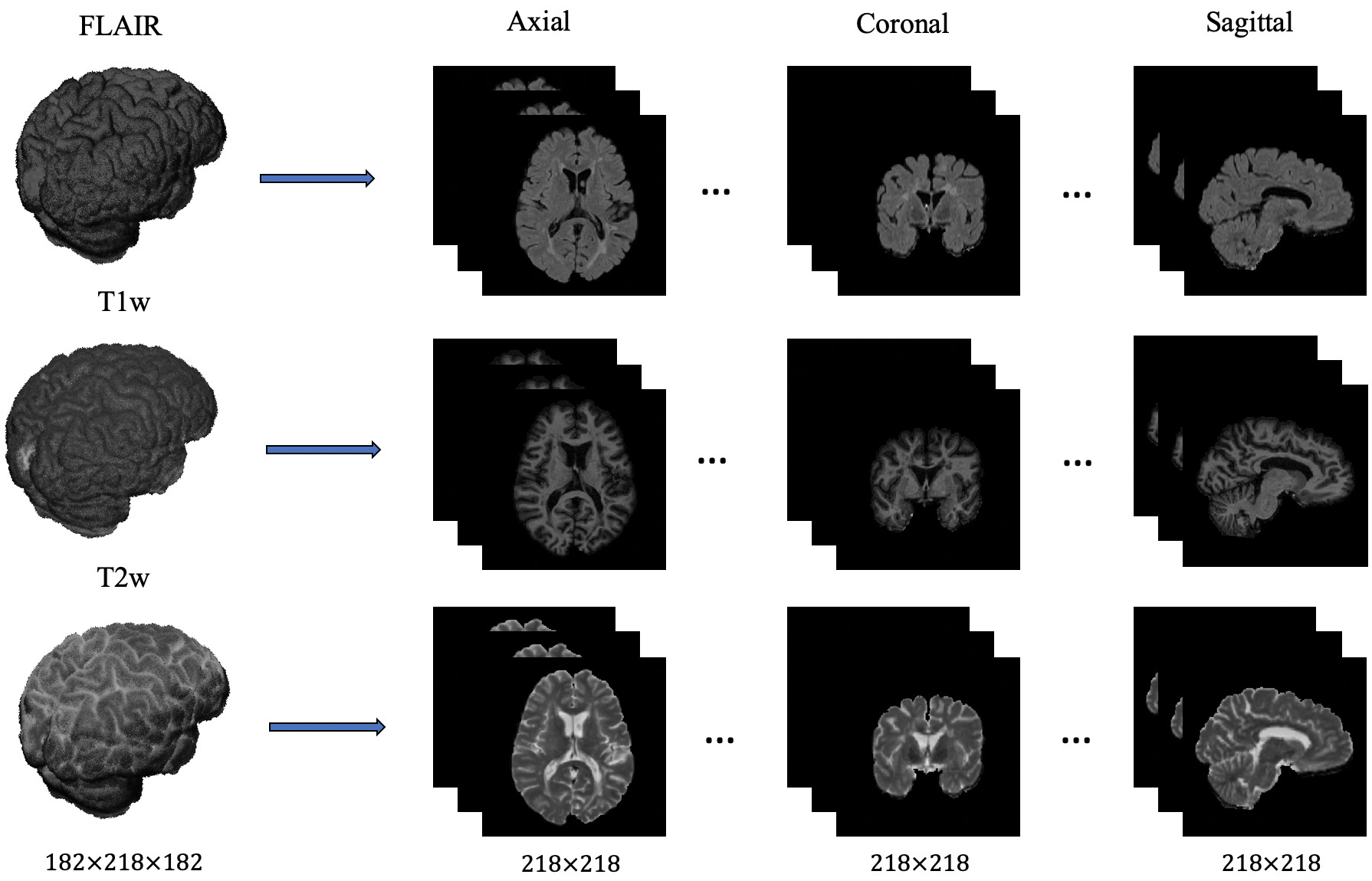}
\caption{Input features preparation. For each subject, three MRI modalities (FLAIR, T1w, and T2w) were considered. 2D slices related to the orthogonal views of the brain (axial, coronal and sagittal planes) were extracted from each modality. Since the size of extracted slices was different with respect to the plane orientations (axial=$182\times218$, coronal=$182\times182$, sagittal=$218\times182$), all slices were zero-padded while centering the brain so to obtain all slices with the same size ($218\times218$), no matter their orientation.} 
\label{fig1}
\end{figure} 

\subsection{Related works}
The literature offers some methods based on CNNs for MS lesion segmentation. For example, \citet{vaidya2015longitudinal} proposed a shallow 3D patch-based CNN using the idea of sparse convolution \citep{li2014highly} for effective training. Moreover, they added a post-processing stage, which increased the segmentation performance by applying a WM mask to the output predictions. \citet{ghafoorian2015convolutional} developed a deep CNN based on 2D patches in order to increase the number of the training samples and avoid the overfitting problems of 3D-based approaches. Similarly, in \citep{birenbaum2016longitudinal}, multiple 2D patch-based CNNs have been designed to take advantage of the common information within longitudinal data. 
\citet{valverde2017improving} proposed a pipeline relying on a cascade of two 3D patch-based CNNs. They trained the first network using all extracted patches, and the second network was used to refine the training procedure utilizing misclassified samples from the first network. \citet{roy2018multiple} proposed a 2D patch-based CNN including two pathways. They used different MRI modalities as input for each pathway and the outputs were concatenated to create a membership function for lesions. Recently, \citet{DBLP:journals/corr/abs-1803-11078} proposed a method relying on a 3D patch-based CNN using the idea of a densely connected network. They also developed an asymmetric loss function for dealing with highly unbalanced data. Despite the fact that all the proposed patch-based techniques have good segmentation performance, they suffer from lacking global structural information. This means that global structure of the brain and the absolute location of lesions are not exploited during the segmentation.

In contrast, \citet{brosch2016deep} developed a whole-brain segmentation method using a 3D CNN. They used single shortcut connection between the coarsest and the finest layers of the network, which enables the network to concatenate the features from the deepest layer to the shallowest layer in order to learn information about the structure and organization of MS lesions. However, they did not exploit middle-level features, which have been shown to have a considerable impact on the segmentation performance \citep{ronneberger2015u}

\subsection{Contributions}
In this paper, we propose a novel deep learning architecture for automatic MS lesion segmentation consisting of a multi-branch 2D convolutional encoder-decoder network. In this study, we concentrated on whole-brain slice-based segmentation in order to prevent both the overfitting present in 3D-based segmentation \citep{brosch2016deep} and the lack of global structure information in patch-based methods \citep{ghafoorian2017deep, valverde2017improving, roy2018multiple}. We designed an end-to-end encoder-decoder network including a multi-branch downsampling path as the encoder, a multi-scale feature fusion and the multi-scale upsampling blocks as the decoder.

In the encoder, each branch is assigned to a specific MRI modality in order to take advantage of each modality individually. During the decoding stage of the network, different scales of the encoded attributes related to each modality, from the coarsest to the finest, including the middle-level attributes, were combined together and upconvolved gradually to get fine details (more contextual information) of the lesion shape. Moreover, we used three different (orthogonal) planes for each 3D modality as an input to the network to better exploit the contextual information in all directions. In summary, the main contributions in this work are:
\begin{itemize}
	\item A whole-brain slice-based approach to exploit the overall structural information, combined with a multi-plane strategy to take advantage of full contextual information.
	\item A multi-level feature fusion and upsampling approach to exploit contextual information at multiple scales.
	\item The evaluation of different versions of the proposed model so as to find the most performant combination of MRI modalities for MS lesion segmentation.
	\item The demonstration of top performance on two different datasets.
\end{itemize}

\section{Material}
In order to evaluate the performance of the proposed method for MS lesion segmentation, two different datasets were used: the publicly available ISBI 2015 Longitudinal MS Lesion Segmentation Challenge dataset \citep{carass2017longitudinal} (denoted as the ISBI dataset), and an in-house dataset from the neuroimaging research unit (NRU) in Milan (denoted as the NRU dataset).

\subsection{ISBI 2015 Longitudinal MS Lesion Segmentation Challenge}
The ISBI dataset included 19 subjects divided into two sets, 5 subjects in the training set and 14 subjects in the test set. Each subject had different time-points, ranging from 4 to 6. For each time-point, T1w, T2w, PDw, and FLAIR image modalities were provided. The volumes were composed of 182 slices with FOV=182$\times$256 and 1-millimeter cubic voxel resolution. All images available were already segmented manually by two different raters, therefore representing two ground truth lesion masks. For all 5 training images, lesion masks were made publicly available. For the remaining 14 subjects in the test set, there was no publicly available ground truth. The performance evaluation of the proposed method over the test dataset was done through an online service by submitting the binary masks to the challenge\footnote{\url{http://iacl.ece.jhu.edu/index.php/MSChallenge}} website \citep{carass2017longitudinal}.

\subsection{Neuroimaging Research Unit}
\label{subsec:NRUdata}

The NRU dataset was collected by a research team from Ospedale San Raffaele, Milan, Italy.

It consisted of 37 MS patients (22 females and 15 males) with mean age $44.6\pm12.2$ years. The patient clinical phenotypes were 24 relapsing remitting MS, 3 primary progressive MS and 10 secondary progressive MS. The mean Expanded Disability status Scale (EDSS) was $3.3\pm2$, the mean disease duration was $13.1\pm8.7$ years and the mean lesion load was $6.2\pm5.7$ ml. The dataset was acquired on a 3.0 Tesla Philips Ingenia CX scanner (Philips Medical Systems) with standardized procedures for subjects positioning.

The following sequences were collected: Sagittal 3D FLAIR sequence, FOV=256$\times$256, pixel size=1$\times$1 mm, 192 slices, 1-mm thick; Sagittal 3D T2w turbo spin echo (TSE) sequence, FOV=256$\times$256, pixel size=1$\times$1 mm, 192 slices, 1-mm thick; Sagittal 3D high resolution T1w, FOV=256$\times$256, pixel size=1$\times$1 mm, 204 slices, 1-mm thick. 

For the validation of the NRU dataset, two different readers, with more than 5 years of experience in manual T2 hyperintense MS lesion segmentation performed the lesion delineation blinded to each other's results. We estimated the agreement between the two expert raters by using the Dice similarity coefficient (\textit{DSC}) as a measure of the degree of overlap between the segmentations, and we found a mean \textit{DSC} of 0.87. Differently from ISBI dataset, the two masks created by the two expert raters were used to generate a high quality ``gold standard'' mask by the intersection of the two binary masks from the two raters, which was used for all experiments with this dataset. This was to follow the common clinical practice of considering a single consensus mask between raters, which was particularly justified in our case due to the high \textit{DSC} value between the two raters.

\subsubsection{Ethical Statement}
Approval was received from the local ethical standards committee on human experimentation; written informed consent was obtained from all subjects prior to study participation.

\begin{figure*}[t!]
\centering
\includegraphics[width=155mm]{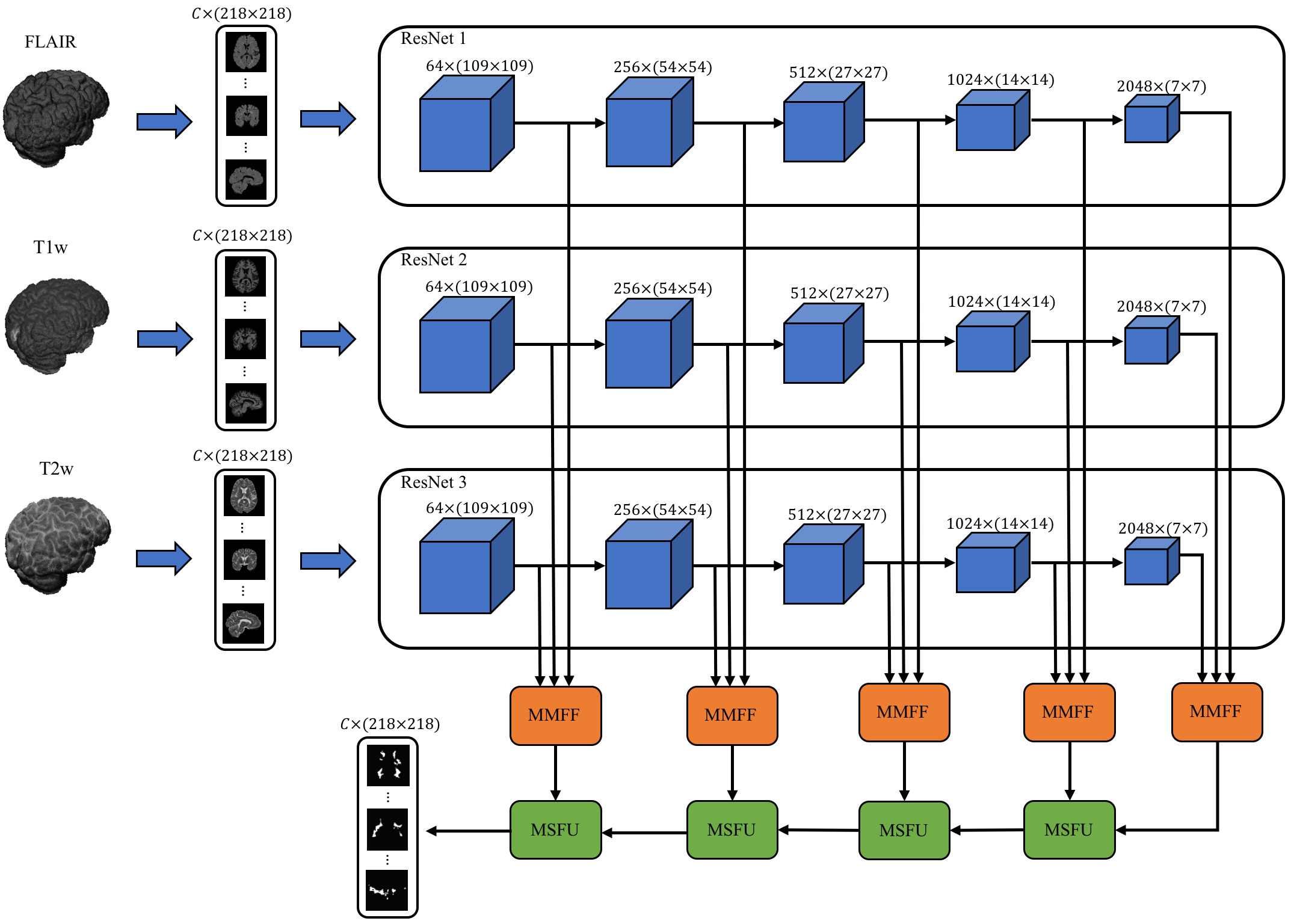}
\caption{General overview of the proposed method. Input data is prepared as described in Section \ref{input}, where volumes for each modality (FLAIR, T1w, and T2w) are described by slices ($C$ is the total number of the slices along axial, coronal, and sagittal orientations, and $218\times218$ is their size after zero-padding). Data is presented in input by slices, and the model generates the corresponding segmented slices. The downsampling part of the network (blue blocks) includes three parallel ResNets without weight sharing, each branch for one modality (in this Figure, we used three modalities: FLAIR, T1w, and T2w). Each ResNet can be considered composed by 5 blocks according to the resolution of the representations. For example, the first block denotes $64$ representations with resolution $109\times109$. Then, MMFF blocks are used to fuse the representations with the same resolution from different modalities. Finally, the output of MMFF blocks is presented as input to MSFU blocks, which are responsible for upsampling the low-resolution representations and for combining them with high-resolution representations.}
\label{fig2}
\end{figure*}

\begin{figure*}[t!]
\centering
\includegraphics[width=145mm]{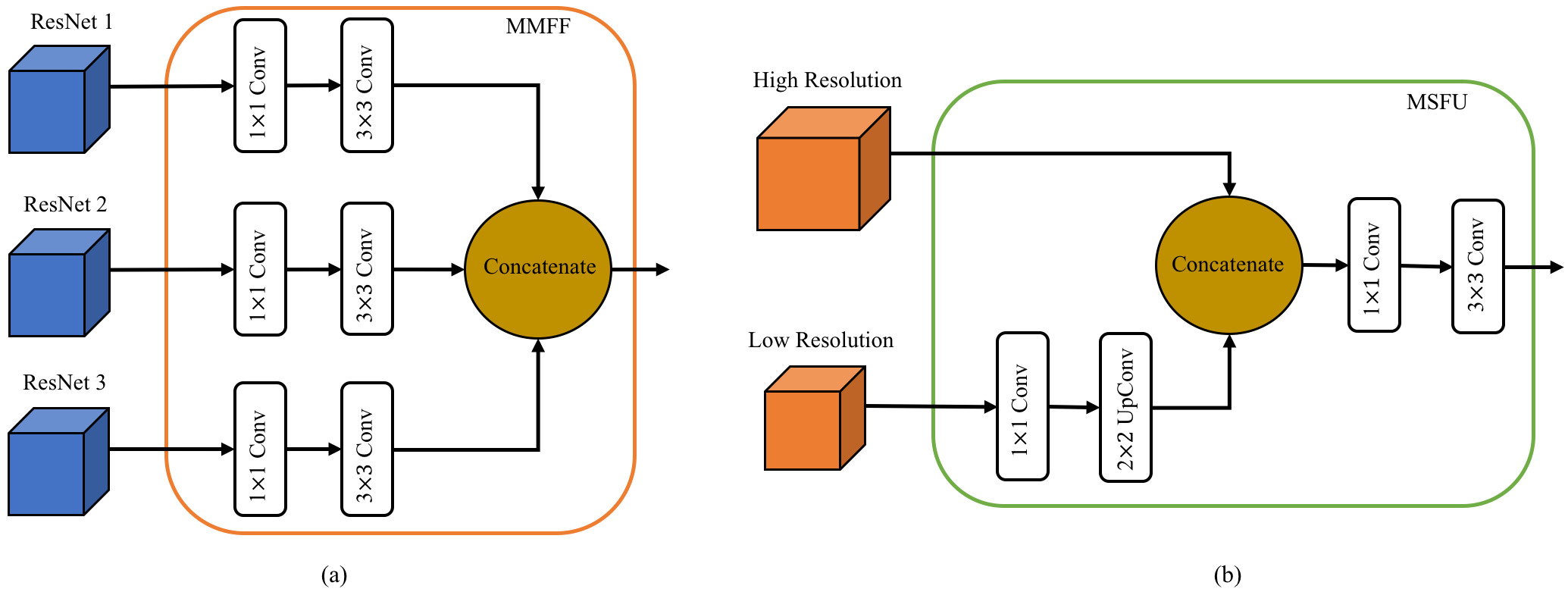}
\caption{Building blocks of the proposed network. a) MMFF block is used to combine representations from different modalities (FLAIR, T1w, and T2w) at the same resolution. b) MSFU block is used to upsample low-resolution features and combine them with higher-resolution features.} 
\label{fig3}
\end{figure*} 

\section{Method}

\subsection{Data Preprocessing}
\label{Data Preprocessing}
From the ISBI dataset, we selected the preprocessed version of the images available online at the challenge website. All images were already skull-stripped using Brain Extraction Tool (BET) \citep{smith2002fast}, rigidly registered to the 1$mm^3$ MNI-ICBM152 template \citep{oishi2008human} using FMRIB's Linear Image Registration tool (FLIRT) \citep{jenkinson2001global, jenkinson2002improved} and N3 intensity normalized \citep{sled1998nonparametric}.

In the NRU dataset, all sagittal acquisitions were reoriented in axial plane and the exceeding portion of the neck was removed. T1w and T2w sequences were realigned to the FLAIR MRI using FLIRT and brain tissues were separated from non-brain tissues using BET on FLAIR volumes. The resulting brain mask was then used on both registered T1w and T2w images to extract brain tissues. Finally, all images were rigidly registered to a 1$mm^3$ MNI-ICBM152 template using FLIRT to obtain volumes of size ($182\times218\times182$) and then N3 intensity normalized. 

\subsection{Network Architecture}
In this work, we propose a 2D end-to-end convolutional network based on the residual network (ResNet) \citep{he2016deep}. The core idea of ResNet is the use of identity shortcut connections, which allows for both preventing gradient vanishing and reducing computational complexity. Thanks to these benefits, ResNets have shown outstanding performance in computer vision problems, specifically in image recognition task \citep{he2016deep}.

We modified ResNet50 (version with 50 layers) to work as a pixel-level segmentation network. This has been obtained by changing the last prediction layer with other blocks and a dense pixel-level prediction layer inspired by the idea of the fully convolutional network (FCN) \citep{long2015fully}. To exploit the MRI multi-modality analysis, we built a pipeline of parallel ResNets without weights sharing. Moreover, a multi-modal feature fusion block (MMFF) and a multi-scale feature upsampling block (MSFU) were proposed to combine and upsample the features from different modalities and different resolutions, respectively.

In the following Sections, we first describe how the input features were generated by decomposing 3D data into 2D images. Then, we describe the proposed network architecture in details and the training procedure. Finally, we introduce the multi-plane reconstruction block, which defines how we combined the 2D binary slices of the network output to match the original 3D data.

\begin{figure*}[t!]
\centering
\includegraphics[width=120mm]{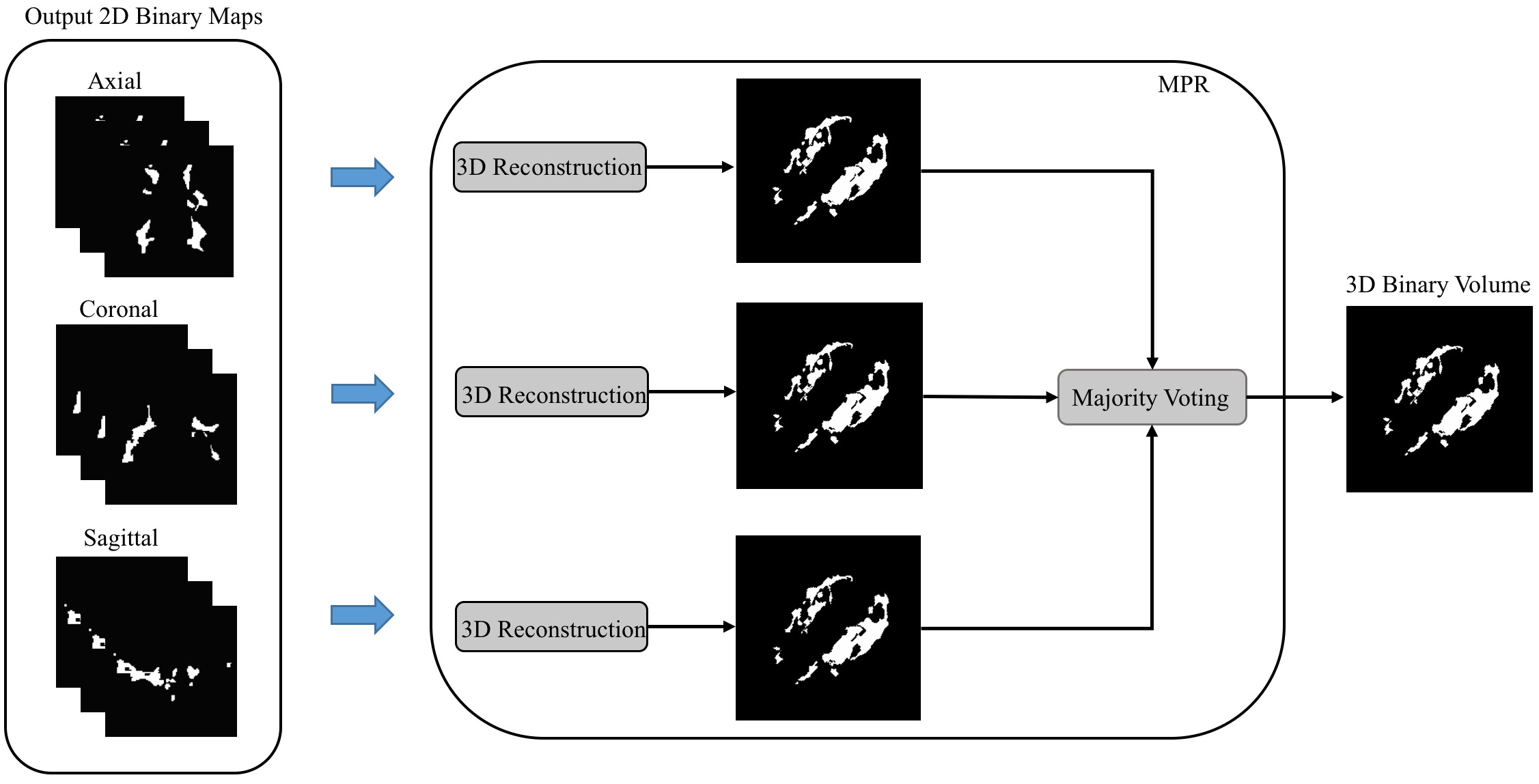}
\caption{The MPR block produces a 3D volumetric binary map by combining the 2D output binary maps of the network. First, the output 2D binary maps associated to each plane orientation (axial, coronal, and sagittal) are concatenated to create three 3D binary maps. Then, a majority vote is applied to obtain a single lesion segmentation volume.} 
\label{fig4}
\end{figure*}

\begin{figure}[b!]
\centering
\includegraphics[width=60mm]{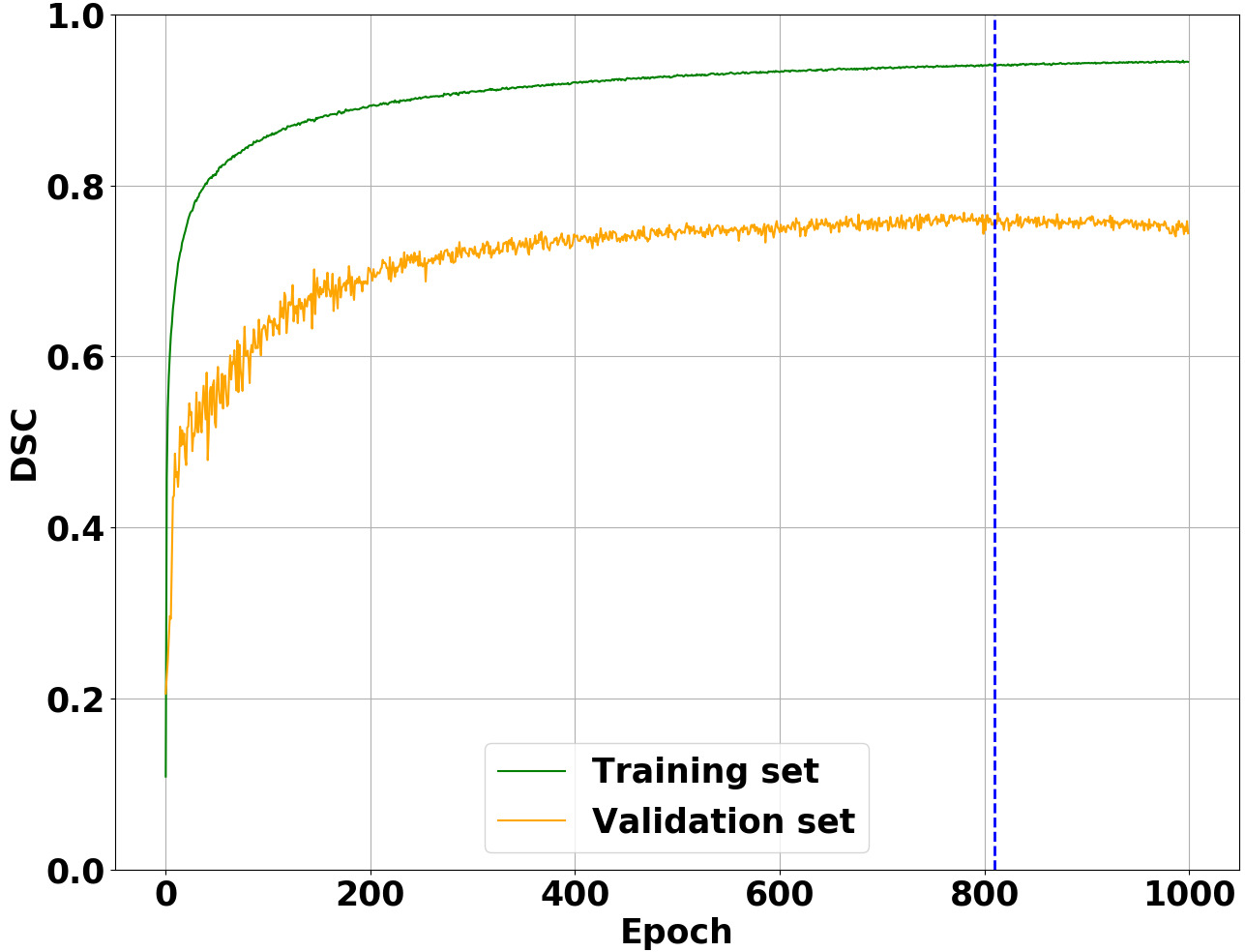}
\caption{Example of \textit{DSC} metric dynamics (eq.~\ref{eq:dsc}) during training on ISBI dataset. Experimentally, we found that a performance plateau was systematically reached before 1000 training epochs. To avoid overfitting, the best model was selected according to the validation set performance. In this specific experiment (training: subjects 1 to 4, validation: subject 5), the best model was selected based at epoch 810, which corresponded to the performance peak on validation set.}
\label{fig6}
\end{figure}

\subsubsection{Input Features Preparation}
\label{input}

For each MRI volume (and each modality), three different plane orientations (axial, coronal and sagittal) were considered in order to generate 2D slices along x, y, and z axes. Since the size of each slice depends on the orientation (axial=$182\times218$, coronal=$182\times182$, sagittal=$218\times182$), they were zero-padded (centering the brain) to obtain equal size ($218\times218$) for each plane orientation. This procedure was applied to all three modalities. \autoref{fig1} illustrates the described procedure using FLAIR, T1w, and T2w modalities. This approach is similar to the one proposed in \citep{roth2014new}, where they used a 2.5D representation of 3D data.

\subsubsection{Network Architecture Details}
\label{network architecture detail}
The proposed model essentially integrates multiple ResNets with other blocks to handle multi-modality and multi-resolution approaches, respectively. As can be seen in \autoref{fig2}, the proposed network includes three main parts: downsampling networks, multi-modal feature fusion using MMFF blocks, and multi-scale upsampling using MSFU blocks.

In the downsampling stage, multiple parallel ResNets (without weights sharing) are used for extracting multi-resolution features, with each ResNet associated to one specific modality (in our experiments, we used FLAIR, T1w, and T2w). In the original ResNet50 architecture, the first layer is composed of a $7 \times 7$ convolutional layer with stride 2 to downsample the input by an order of 2. Then, a $3 \times 3$ max pooling layer with stride 2 is applied to further downsample the input followed by a bottleneck block without downsampling. Subsequently, three other bottleneck blocks are applied, each one followed by a downsampling convolutional layer with stride $2$.

Therefore, ResNet50 can be organized into five blocks according to the resolution of the generated feature maps ($109\times109$, $54\times54$, $27\times27$, $14\times14$, and $7\times7$). Thanks to this organization, we can take advantage of the multi-resolution.
Features with the same resolution from different modalities are combined using MMFF blocks as illustrated in \autoref{fig3}(a). Each MMFF block includes $1 \times 1$ convolutions to reduce the number of feature maps (halving them), followed by $3\times3$ convolutions for adaptation. A simple concatenation layer is then used to combine the features from different modalities. 

In the upsampling stage, MSFU blocks fuse the multi-resolution representations and gradually upsize them back to the original resolution of the input image. \autoref{fig3}(b) illustrates the proposed MSFU block consisting of a $1 \times1$ convolutional layer to reduce the number of feature maps (halving them) and an upconvolutional layer with $2 \times 2$ kernel size and a stride of 2, transforming low-resolution feature maps to higher resolution maps. Then, a concatenation layer is used to combine the two sets of feature maps, followed by a $1 \times1$ convolutional layer to reduce the number of feature maps (halving them) and a $3\times3$ convolutional layer for adaptation. 

After the last MSFU block, a soft-max layer of size 2 is used to generate the output probability maps of the lesions. In our experiments the probabilistic maps were thresholded at 0.5 to generate binary classification for each pixel (lesion vs. non-lesion). It is important to mention that in all proposed blocks before each convolutional and upconvolutional layer, we use a batch normalization layer \citep{ioffe2015batch} followed by a rectifier linear unit activation function \citep{nair2010rectified}. Size and number of feature maps in the input and output of all convolutional layers are kept the same.

\subsubsection{Implementation Details}
\label{Implementation Details}
The proposed model was implemented in Python language\footnote{\url{https://www.python.org}} using Keras\footnote{\url{https://keras.io}} \citep{chollet2015keras} with Tensorflow\footnote{\url{https://www.tensorflow.org}} \citep{tensorflow2015-whitepaper} backend. All experiments were done on a Nvidia GTX Titan X GPU. Our multi-branch slice-based network was trained end-to-end. In order to train the proposed CNN, we created a training set including the 2D slices from all three orthogonal views of the brain, as described in Section \ref{input}. Then, to limit extremely unbalanced data and omit uninformative samples, a training subset was determined by selecting only slices containing at least one pixel labeled as lesion. Considering that for each subject in the ISBI dataset, there were 4 to 6 recordings, the number of slices selected per subject ranged approximately from 1500 to 2000. In the NRU dataset, the number of slices ranged approximately from 150 to 300 per subject.

To optimize the network weights and early stopping criterion, the created training set was divided into training, and validation subsets, depending on the experiments described in the following Section (In all experiments, the split was performed on the subject base, to simulate a real clinical condition).
We trained our network using the Adam optimizer \citep{DBLP:journals/corr/KingmaB14} with an initial learning rate of 0.0001 multiplied by 0.95 every 400 steps. The size of mini-batches was fixed at 15 and each mini-batch included random slices from different orthogonal views. The maximum number of training epochs was fixed to 1000 for all experiments, well beyond the average converging rate. \autoref{fig6} illustrates an example of performance evolution during training of the network in terms of mean \textit{DSC} (refer to \ref{metrics} for details). Indeed, a performance plateau was systematically observed over all experiments before 1000 epochs. The best model was then selected according to the validation set. In the case shown on \autoref{fig6}, the best performance was obtained at epoch 810. The training computation time for 1000 epochs was approximately 36 hours.

Regarding the network initialization, in the downsampling branches, we used ResNet50 pre-trained on ImageNet and all other blocks (MMFFs and MSFUs) were randomly initialized from a Gaussian distribution with zero mean and standard deviation equal to $\sqrt{2/(a+b)}$ where $a$ and $b$ are respectively the number of input and output units in the weight tensor. It is worth noticing that we did not use parameter sharing in parallel ResNets. The soft Dice Loss function (DL) was used to train the proposed network:

    \begin{equation}
        \label{eq:DL}
        \textit{DL} =1-\frac{2 \sum_{i}^{N} g_i p_i} {\sum_{i}^{N} {g_i}^2 + \sum_{i}^{N} {p_i}^2}
    \end{equation}

where $p_i\in[0,...,1]$ is the predicted value of the soft-max layer and $g_i$ is the ground truth binary value for each pixel $i$.

We slightly modified the original soft dice loss \citep{milletari2016v} by replacing (-Dice) with (1-Dice) for visualization purposes. Indeed, the new equation returns positive values in the range $[0,...,1]$. This change does not impact the optimization.

\subsubsection{3D Binary Image Reconstruction}
\label{reconstruction} 
Output binary slices of the network are concatenated to form a 3D volume matching the original data. In order to reconstruct the 3D image from the output binary 2D slices, we proposed a multi-planes reconstruction (MPR) block. Feeding each 2D slice to the network, we get as output the associated 2D binary lesion classification map. Since each original modality is duplicated three times in the input, once for each slice orientation (coronal, axial, sagittal), concatenating the binary lesion maps belonging to the same orientation results in three 3D lesion classification maps. To obtain a single lesion segmentation volume, these three lesion maps are combined via majority voting (the most frequent lesion classification are selected) as illustrated in \autoref{fig4}. To justify the choice of majority voting instead of other label fusion methods, refer to \ref{app:B}.

\subsection{Data and Code Availability Statement}
The NRU dataset is a private clinical dataset and can not be made publicly available due to confidentiality. The code will be made available to anyone contacting the corresponding  authors.

\section{Experiments}
\subsection{Evaluation Metrics}
\label{metrics}
The following measures were used to evaluate and compare our model with other state-of-the-art methods.  
\begin{itemize}
	\item Dice Similarity Coefficient:
	\begin{equation}
	   \label{eq:dsc}
	    \textit{DSC}=\frac{\textit{2TP}}{\textit{FN + FP + 2TP}}
    \end{equation}
    where \textit{TP}, \textit{FN} and \textit{FP} indicate the true positive, false negative and false positive voxels, respectively.

    \item Lesion-wise True Positive Rate:
    \begin{equation}
    	\textit{LTPR}=\frac{\textit{LTP}}{\textit{RL}}
    \end{equation}
    where \textit{LTP} denotes the number of lesions in the reference segmentation that overlap with a lesion in the output segmentation (at least one voxel overlap), and \textit{RL} is the total number of lesions in the reference segmentation.

    \item Lesion-wise False Positive Rate:
    \begin{equation}
    	\textit{LFPR}=\frac{\textit{LFP}}{\textit{PL}}
    \end{equation}
    where \textit{LFP} denotes the number of lesions in the output segmentation that do not overlap with a lesion in the reference segmentation and \textit{PL} is the total number of lesions in the produced segmentation.
    
    \item Average Symmetric Surface Distance:
    \begin{equation}
      \textit{SD}=\frac{1}{|N_{gt}| + |N_{s}|} \cdot \left(\sum_{x \in {N_{gt}}}\min_{y \in {N_{s}}} d(x,y) + \sum_{x \in {N_{s}}} \min_{y \in {N_{gt}}}d(x,y)\right)
    \end{equation}  
    where $N_{s}$ and $N_{gt}$ are the set of voxels in the contour of the automatic and manual annotation masks, respectively. $d(x,y)$ is the Euclidean distance (quantified in millimetres) between voxel $x$ and $y$.   
     
     \item Hausdorff Distance:
     \begin{equation}
   	\textit{HD}=\max\left\{\max_{x \in {N_{gt}}} \min_{y \in {N_{s}}} d(x,y), \max_{x \in {N_{s}}} \min_{y \in {N_{gt}}} d(x,y) \right\}
    \end{equation} 
    \end{itemize}
    
As described in \citep{carass2017longitudinal}, the ISBI challenge website provides a report on the submitted test set including some measures such as:
\begin{itemize}
\item Positive Prediction Value:
    \begin{equation}
    	\textit{PPV}=\frac{\textit{TP}}{\textit{TP + FP}}
    \end{equation}   
\item Absolute Volume Difference:
    \begin{equation}
   	\textit{VD}=\frac{|\textit{TP}_s-\textit{TP}_{gt}|}{\textit{TP}_{gt}}
    \end{equation}
    where $\textit{TP}_{s}$ and $\textit{TP}_{gt}$ reveal the total number of the segmented lesion voxels in the output and manual annotations masks, respectively. 
 \item Overall evaluation score:
   \begin{equation}
 	    \small
 		\textit{SC} =\frac{1}{|R|\cdot |S|} \cdot \sum_{R,S}\left(\frac{\textit{DSC}}{8}+\frac{\textit{PPV}}{8}+\frac{1-\textit{LFPR}}{4}+\frac{\textit{LTPR}}{4}+\frac{\textit{Cor}}{4}\right)
 	\end{equation}
 	where \textit{S} is the set of all subjects, \textit{R} is the set of all raters and \textit{Cor} is the Pearson's correlation coefficient of the volumes.
\end{itemize}  

\subsection{Experiments on the ISBI Dataset}
\label{Experiments on the ISBI dataset)}
To evaluate the performance of the proposed method on the ISBI dataset, two different experiments were performed according to the availability of the ground truth.

Since the ground truth was available only for the training set, in the first experiment, we ignored the official ISBI test set. We only considered data with available ground truth (training set with 5 subjects) as mentioned in \citep{brosch2016deep}. To obtain a fair result, we tested our approach with a nested leave-one-subject-out cross-validation (3 subjects for training, 1 subject for validation and 1 subject for testing - refer to \ref{app:A} for more details).
To evaluate the stability of the model, this experiment was performed evaluating separately our method on the two sets of masks provided by the two raters.

In the second experiment, the performance of the proposed method was evaluated on the official ISBI test set (with 14 subjects), for which the ground truth was not available, using the challenge web service. We trained our model doing a leave-one-subject-out cross-validation on the whole training set with 5 subjects (4 subjects for training and 1 subject for validation - refer to \ref{app:A} for more details). We executed the ensemble of 5 trained models on the official ISBI test set and the final prediction was generated with a majority voting over the ensemble. The 3D output binary lesion maps were then submitted to the challenge website for evaluation.

\begin{table*}[t!]
\scriptsize
\centering
\caption{Comparison of our method with other state-of-the-art methods in the first ISBI dataset experiment (in this experiment, only images with available ground truth were considered). GT1 and GT2 denote the corresponding model was trained using annotation provided by rater 1 and rater 2 as  ground truth, respectively (the model was trained using GT1 and tested using both GT1 and GT2 and vice versa). Mean values of \textit{DSC}, \textit{LTPR}, and \textit{LFPR} for different methods are shown. Values in bold and italic refer to the first-best and second-best values of the corresponding metrics, respectively.}
\begin{tabular}{lcccccc}
\hline
\multicolumn{1}{c}{Method} & & Rater 1 & & & Rater 2 & \\
                           & DSC & LTPR & LFPR & DSC & LTPR & LFPR \\ \hline
Rater 1 & - & - & - & 0.7320 & 0.6450 & 0.1740  \\
Rater 2 & 0.7320 & 0.8260 & 0.3550 & - & - & -  \\
\citet{maier2015ms} (GT1) & \textit{0.7000} & 0.5333 & 0.4888 & \textit{0.6555} & 0.3777 & \textit{0.4444}                \\
\citet{maier2015ms} (GT2) & \textit{0.7000} & 0.5555 & \textit{0.4888} & 0.6555 & 0.3888 & \textit{0.4333}                \\
\citet{brosch2016deep} (GT1) & 0.6844 & \textit{0.7455} & 0.5455 & 0.6444 & \textit{0.6333} & 0.5288 \\
\citet{brosch2016deep} (GT2) & 0.6833 & \textit{0.7833} & 0.6455 & 0.6588 & \textit{0.6933} & 0.6199 \\
\citet{Aslani_2019} (GT1) & 0.6980 &\textbf{0.7460} & \textit{0.4820} & 0.6510 & \textbf{0.6410} & 0.4506 \\ 
\citet{Aslani_2019} (GT2) & 0.6940 & \textbf{0.7840} & 0.4970 & \textit{0.6640} & \textbf{0.6950} & 0.4420 \\ 
Ours (GT1) & \textbf{0.7649} & 0.6697 & \textbf{0.1202} & \textbf{0.6989} & 0.5356 & \textbf{0.1227}  \\ 
Ours (GT2) & \textbf{0.7646} & 0.7002 & \textbf{0.2022} & \textbf{0.7128} & 0.5723 & \textbf{0.1896}                \\ 
\hline
\end{tabular}
\label{table1}
\end{table*}

\begin{table*}[t!]
\scriptsize
\centering
\caption{Results related to the top-ranked methods (with published papers or technical reports) evaluated on the official ISBI test set and reported on the ISBI challenge website. \textit{SC}, \textit{DSC}, \textit{PPV}, \textit{LTPR}, \textit{LFPR}, and \textit{VD} are mean values across the raters. For detailed information about the metrics, refer to Section \ref{metrics}. Values in bold and italic refer to the metrics with the first-best and second-best performances, respectively.}
\begin{tabular}{lcccccc}
\hline
\multicolumn{1}{c}{Method} &SC &DSC &PPV &LTPR &LFPR &VD \\ \hline

\citet{DBLP:journals/corr/abs-1803-11078} &\textbf{92.48} &0.5841 &\textbf{0.9207} &0.4135 &\textbf{0.0866} &0.4972 \\

Ours & \textit{92.12} &0.6114 &\textit{0.8992} &0.4103 &\textit{0.1393} &0.4537 \\  

\citet{andermatt2017automated} &92.07 &\textit{0.6298} &0.8446 &0.4870 &0.2013 &0.4045 \\

\citet{valverde2017improving} &91.33 &\textbf{0.6304} &0.7866 &0.3669 &0.1529 &\textbf{0.3384}  \\ 

\citet{maier2015ms} &90.28 &0.6050 &0.7746 &0.3672 &0.2657 &0.3653  \\

\citet{birenbaum2016longitudinal} &90.07 &0.6271 &0.7889 &\textbf{0.5678} &0.4975 &\textit{0.3522} \\

\citet{Aslani_2019} &89.85 &0.4864 &0.7402 &0.3034 &0.1708 &0.4768 \\ 

\citet{deshpande2015adaptive} &89.81 &0.5960 &0.7348 &0.4083 &0.3075 & 0.3762 \\

\citet{jain2015automatic} &88.74 &0.5560 &0.7300 &0.3225 &0.3742 &0.3746 \\

\citet{sudre2015bayesian} &87.38 &0.5226 &0.6690 &\textit{0.4941} &0.6776 & 0.3837\\

\citet{tomas2015model} &87.01 &0.4317 &0.6973 &0.2101 &0.4115 &0.5109  \\

\citet{ghafoorian2017deep} &86.92 &0.5009 &0.5491 &0.4288 &0.5765 &0.5707 \\
\hline
\end{tabular}
\label{table2}
\end{table*}

\subsection{Experiment on the NRU Dataset}
\label{Experiment on the NRU dataset}

To test the robustness of the proposed model, we performed two experiments using the NRU dataset including 37 subjects.
In the first experiment, we implemented a nested 4-fold cross-validation over the whole dataset (21 subjects for training, 7 subjects for validation and 9 subjects for testing - refer to \ref{app:A} for more details). Since for each test fold we had an ensemble of four nested trained models, the prediction on each test fold was obtained as a majority vote of the corresponding ensemble.
To justify the use of majority voting instead of other label fusion methods, we repeated the same experiment using different volume aggregation methods (refer to \ref{app:B} for more details).

For comparison, we tested three different publicly available MS lesion segmentation software: OASIS (Automated Statistic Inference for Segmentation) \citep{sweeney2013oasis}, TOADS (Topology reserving Anatomy Driven Segmentation) \citep{shiee2010topology}, and LST (Lesion Segmentation Toolbox)\citep{schmidt2012automated}.
OASIS generates the segmentation exploiting information from FLAIR, T1w, and T2w modalities, and it only requires a single thresholding parameter, which was optimized to obtain the best \textit{DSC}. TOADS does not need parameter tuning and it only requires FLAIR and T1w modalities for segmentation. Similarly, LST works with FLAIR and T1w modalities only. However, it needs a single thresholding parameter that initializes the lesion segmentation. This parameter was  optimized to get the best \textit{DSC} in this experiment.

We also tested the standard 2D U-Net \citep{ronneberger2015u}, repeating the training protocol described in \ref{app:A}. Indeed, we used the same training set as described in Section \ref{input} and \ref{Implementation Details}, with the difference that 2D slices from all modalities were aggregated in multiple channels. This network was trained using the Adam optimizer \citep{DBLP:journals/corr/KingmaB14} with an initial learning rate of 0.0001 multiplied by 0.9 every 800 steps. For the sake of comparison, optimization was performed on the soft Dice Loss function (eq.~\ref{eq:DL}) \citep{milletari2016v}. To get the 3D volume from output binary slices of the network, we used the proposed MPR block as described in Section \ref{reconstruction}.

Differences in performance metrics between our method and each of the 4 other methods were statistically evaluated with resampling. For a given method M and metric C, resampling was performed by randomly assigning for each subject the sign of the difference in C between method M and our method in 10 million samples. The test was two-sided and corrected for multiple comparisons with Holm's method (28 comparisons in total with 7 metrics assessed for the 4 methods to compare ours with). The alpha significance threshold level was set to 0.05.

As outlined in Section~\ref{subsec:NRUdata}, while for the ISBI dataset, we evaluated our method on two separate sets of masks, one for each rater, in the NRU dataset, we considered the manual consensus segmentation as a more robust gold standard against which to validate the proposed method. Nevertheless, to evaluate the stability of the model trained with the gold standard labeling, we also tested it separately on the two sets of masks (refer to \ref{app:C} for more details).

In the second experiment, to investigate the importance of each single modality in MS lesion segmentation, we evaluated our model with various combinations of modalities. This means that the model was adapted in the number of parallel branches in the downsampling network. In this experiment, we randomly split the corresponding dataset into fixed training (21 subjects), validation (7 subjects) and test (9 subjects) sets.

\begin{table*}[t]
\scriptsize
\centering
\caption{Results related to the first NRU dataset experiment. Mean values of \textit{DSC}, \textit{PPV}, \textit{LTPR}, \textit{LFPR}, \textit{VD}, \textit{SD} and \textit{HD} were measured for different methods. Values in bold and italic indicate the first-best and second-best results.}
\begin{tabular}{lccccccc}
\hline
\multicolumn{1}{c}{Method} &DSC &PPV &LTPR &LFPR &VD &SD &HD \\ \hline
TOADS \citep{shiee2010topology} &0.5241 &0.5965 &\textbf{0.4608} &0.6277 &\textit{0.4659} &5.4392 &13.60 \\
LST \citep{schmidt2012automated} &0.3022 &0.5193 &0.1460 &0.3844 &0.6966 &7.0919 &14.35 \\
OASIS \citep{sweeney2013oasis} &0.4193 &0.3483 &0.3755 &0.4143 &2.0588 &\textit{3.5888} &18.33\\
U-NET \citep{ronneberger2015u} &\textit{0.6316} &\textit{0.7748} &0.3091 &\textit{0.2267} &\textit{0.3486} &3.9373 &\textit{9.235}\\
OURS &\textbf{0.6655} &\textbf{0.8032} &\textit{0.4465} &\textbf{0.0842} &\textbf{0.3372} &\textbf{2.5751} &\textbf{6.728}\\
\hline
\end{tabular}
\label{table3}
\end{table*}

\begin{table*}[t!]
\scriptsize
\centering
\caption{The proposed model was tested with different combinations of the three modalities in the second NRU dataset experiment. SB and MB denote the single-branch and multi-branch versions of the proposed model, respectively. Mean values of \textit{DSC}, \textit{PPV}, \textit{LTPR}, \textit{LFPR}, \textit{VD}, \textit{SD} and \textit{HD} were measured for different methods. Values in bold and italic indicate the first-best and second-best values.}
\begin{tabular}{llccccccc}
\hline
\multicolumn{1}{c}{Method} &Set of Modalities &DSC &PPV &LTPR &LFPR &VD &SD &HD \\ \hline
SB & FLAIR & 0.6531 & 0.5995 & 0.6037 & 0.2090 & 0.3034 &1.892 &9.815 \\
   & T1w & 0.5143 & 0.5994 & 0.3769 & 0.2738 & 0.3077 &4.956 &\textit{8.201} \\
   & T2w & 0.5672 & 0.5898 & 0.4204 & 0.2735 & \textit{0.1598} &4.733 &9.389 \\ 
   & FLAIR, T1w, T2w & \textit{0.6712} & 0.6029 & 0.6095 & 0.2080 & 0.2944 &\textit{1.602} &9.989 \\ \hline
MB & FLAIR, T1w & 0.6624 & \textit{0.6109} & \textit{0.6235} & 0.2102 & 0.2740 &1.727 &9.526 \\
   & FLAIR, T2w & 0.6630 & 0.6021 & \textbf{0.6511} & \textit{0.2073} & 0.3093 &1.705 &9.622 \\
   & T1w, T2w & 0.5929 & 0.6102 & 0.4623 & 0.2309 & 0.1960 &4.408 &9.004 \\
   & FLAIR, T1w, T2w & \textbf{0.7067} & \textbf{0.6844} & 0.6136 & \textbf{0.1284} & \textbf{0.1488} &\textbf{1.577} &\textbf{8.368} \\
\hline
\end{tabular}
\label{table4}
\end{table*}

\textbf{Single-branch (SB):} In a single-branch version of the proposed model, we used a single ResNet as the downsampling part of the network. Attributes from different levels of the single-branch were supplied to the MMFF blocks. In this version of our model, each MMFF block had single input since there was only one downsampling branch. Therefore, MMFF blocks included a $1\times1$ convolutional layer followed by a $3\times3$ convolutional layer. We trained and tested the single-branch version of our proposed network with each modality separately and also with a combination of all modalities as a multi-channel input.

\textbf{Multi-branch (MB):} The multi-branch version of the proposed model used multiple parallel ResNets in the downsampling network without weights sharing. In this experiment, we used two-branch and three-branch versions, which were trained and tested using two modalities and three modalities, respectively. We trained and tested the mentioned models with all possible combination of modalities (two-branches: [FLAIR, T1w], [FLAIR, T2w], [T1w, T2w] and three-branches: [FLAIR, T1w, T2w]).

\section{Results}
\subsection{ISBI Dataset}
\label{ISBI}
\begin{figure*}[t!]
\centering
\includegraphics[width=185mm]{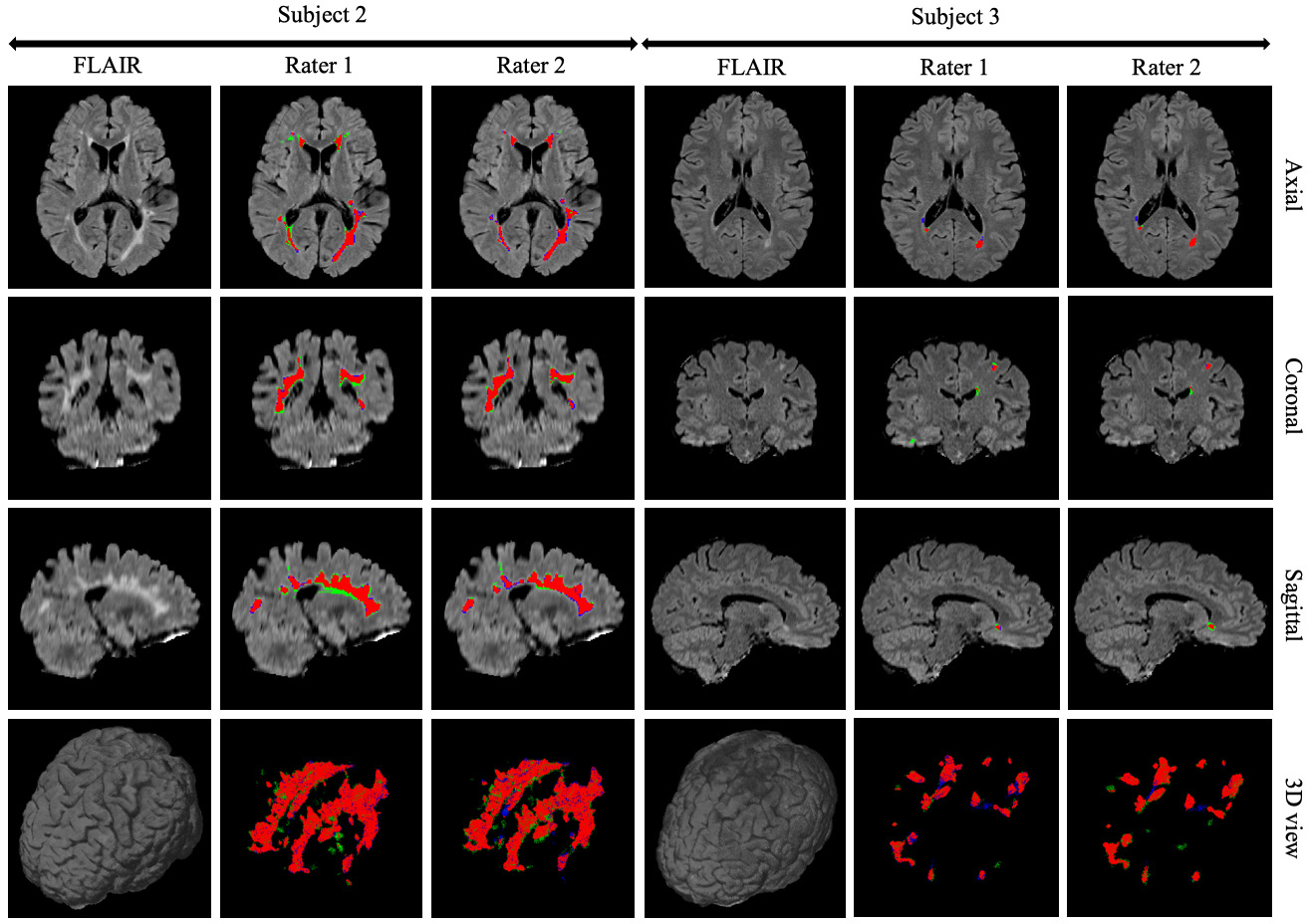}
\caption{Output segmentation results of the proposed method on two subjects of the ISBI dataset compared to ground truth annotations provided by rater 1 and rater 2. From left to right, the first three columns are related to subject 2 with high lesion load and reported \textit{DSC} values of 0.8135 and 0.8555 for rater 1 and rater 2, respectively. Columns 4 to 6 are related to the subject 3 with low lesion load and reported \textit{DSC} values of 0.7739 and 0.7644 for rater 1 and rater 2, respectively. On all images, true positives, false negatives, and false positives are colored in red, green and blue, respectively.} 
\label{fig5}
\end{figure*}

In the first experiment, we evaluated our model using three measures: \textit{DSC}, \textit{LTPR}, and \textit{LFPR} to make our results comparable to those obtained in \citep{brosch2016deep,maier2015ms,Aslani_2019}. \autoref{table1} summarizes the results of the first experiment when comparing our model with previously proposed methods. The table shows the mean \textit{DSC}, \textit{LTPR}, and \textit{LFPR}. As can be seen in that table, our method outperformed other methods in terms of \textit{DSC} and \textit{LFPR}, while the highest \textit{LTPR} was achieved by our recently published method \citep{Aslani_2019}. \autoref{fig5} shows the segmentation outputs of the proposed method for subject 2 (with high lesion load) and subject 3 (with low lesion load) compared to both ground truth annotations (rater 1 and rater 2).

In the second experiment, the official ISBI test set was used. Indeed, all 3D binary output masks computed on the test set were submitted to the ISBI website. Several measures were calculated online by the challenge website. \autoref{table2} shows the results on all measures reported as a mean across raters. At the time of the submission, our method had an overall evaluation score of 92.12 on the official ISBI challenge web service\footnote{\url{http://iacl.ece.jhu.edu/index.php/MSChallenge}}, making it amongst the top-ranked methods with a published paper or a technical report.
 
\subsection{NRU Dataset}  
\autoref{table3} reports the results of the first experiment on NRU dataset showing the mean values of \textit{DSC}, \textit{LFPR}, \textit{LTPR}, \textit{PPV}, \textit{VD}, \textit{SD} and \textit{HD}. It summarizes how our method performed compared to others. As shown in the table, our method achieved the best results with respect to \textit{DSC}, \textit{PPV}, \textit{LFPR}, \textit{VD}, \textit{SD} and \textit{HD} measures while showing a good trade-off between \textit{LTPR} and \textit{LFPR}, comparable to the best results of the other methods.

\begin{figure*}[t!]
\centering
\includegraphics[width=170 mm]{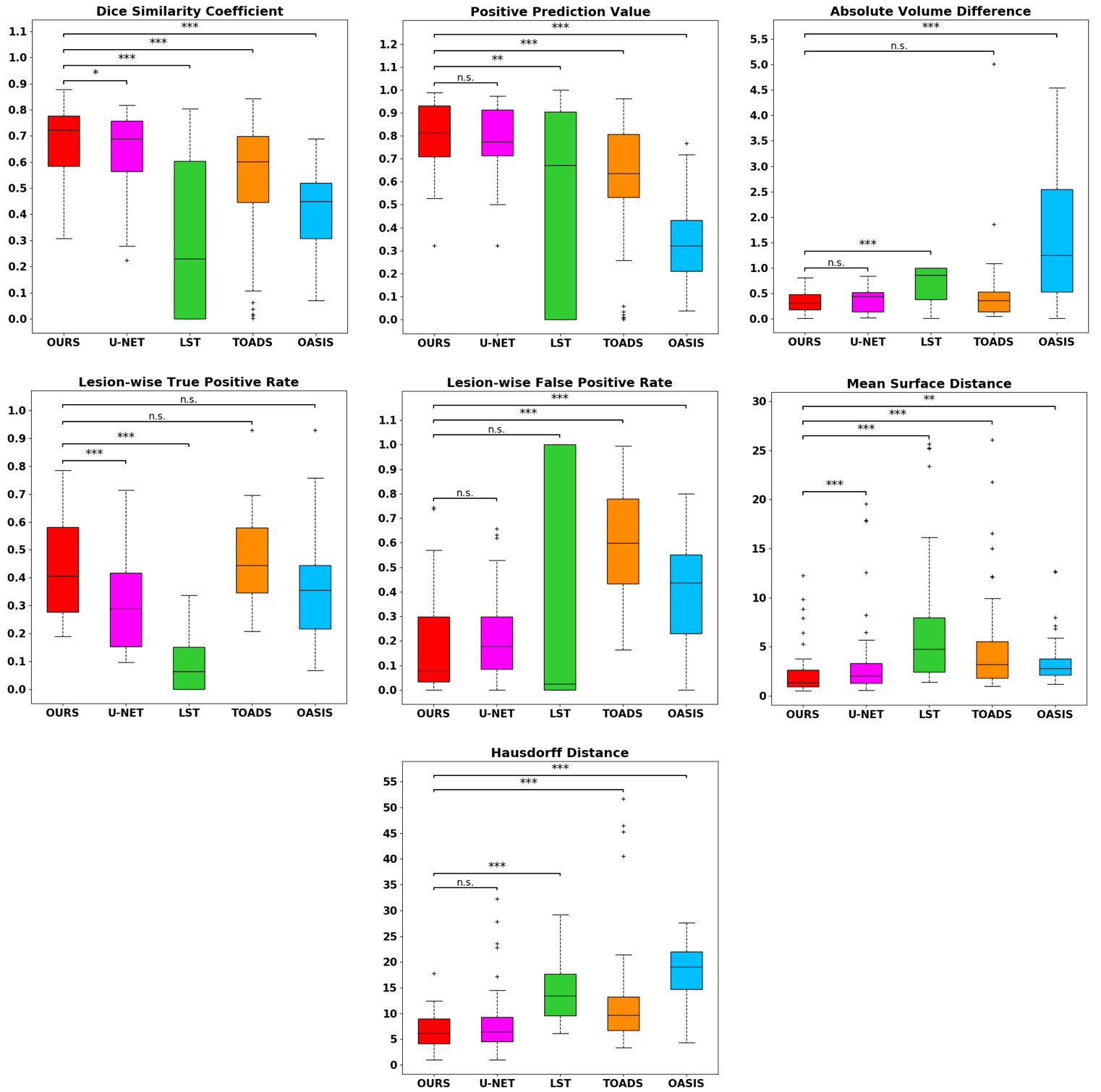}
\caption{Boxplots showing the performance of tested models with all measures on NRU dataset. Among all methods, the proposed one had the best trade-off between the lesion-wise true positive rate and lesion-wise false positive rate, while having the best mean value for dice similarity coefficient, positive prediction value, absolute volume differences, mean surface distance and hausdorff distance. Statistically significant differences between our method and the others were assessed using resampling statistics with multiple comparison correction. The significance threshold was set as $\alpha=0.05$. $p$-values were annotated as follows: '*' for $p<0.05$, '**' for $p<0.005$, '***' for $p<0.0005$, and 'n.s.' for non-significant values.}
\label{fig7}
\end{figure*}

\begin{figure*}[t!]
\centering
\includegraphics[width=185mm]{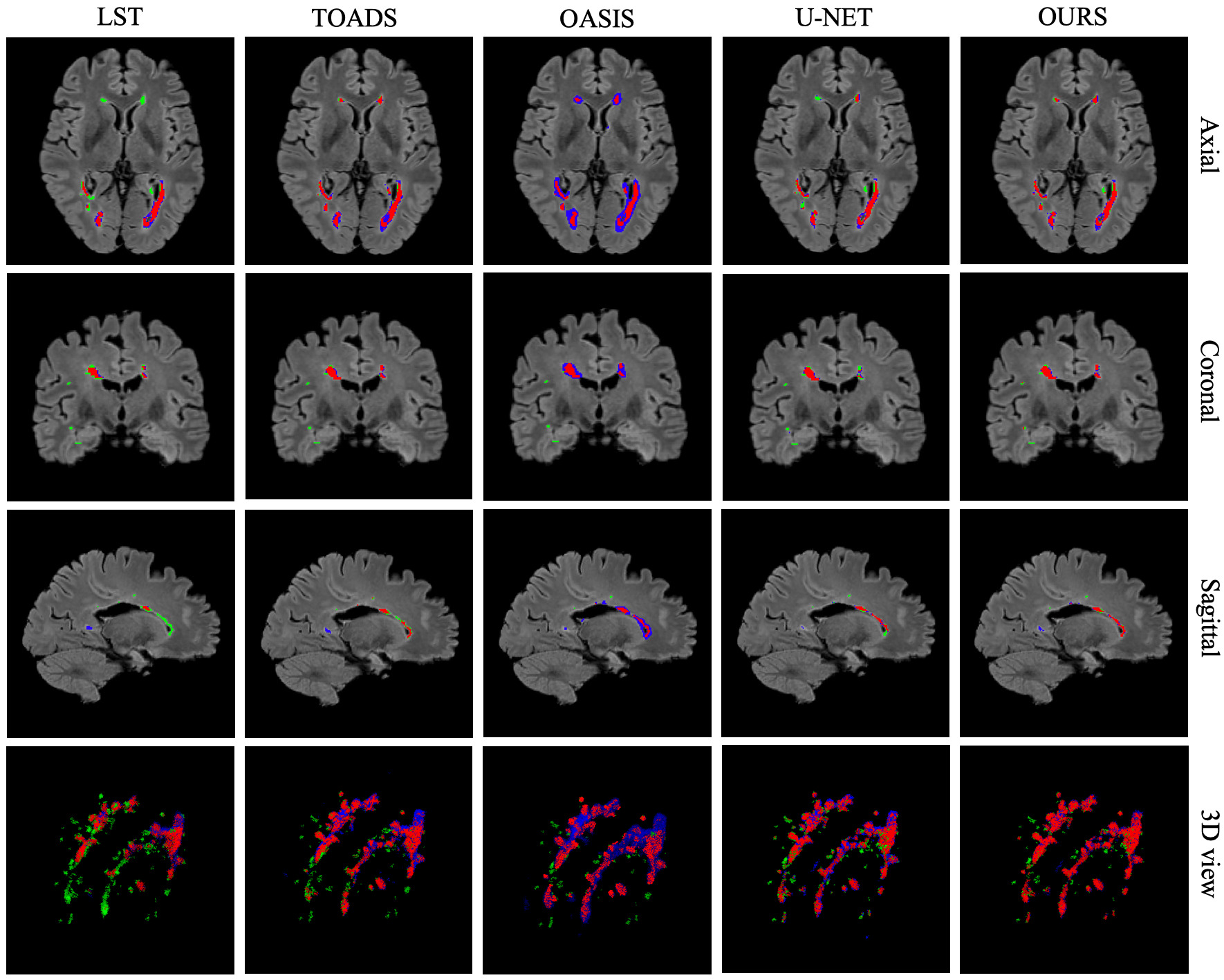}
\caption{Output segmentation results of the different methods for one subject with medium lesion load from the NRU dataset compared with ground truth annotation. Reported \textit{DSC} values for TOADS, OASIS, LST, U-Net and our proposed method for this subject are 0.7110, 0.4266, 0.6505, 0.7290 and 0.7759, respectively. On all images, true positives, false negatives, and false positives are colored in red, green and blue, respectively.} 
\label{fig8}
\end{figure*}

\autoref{fig7} features boxplots of the \textit{DSC}, \textit{LFPR}, \textit{LTPR}, \textit{PPV}, \textit{VD}, \textit{SD} and \textit{HD} evaluation metrics obtained from the different methods and summarized in \autoref{table3}. This Figure shows statistically significant differences between model performances for most metrics and methods when compared to ours, after multiple comparison correction with the conservative Holm's method. The output segmentation of all methods applied to a random subject (with medium lesion load) can be seen with different plane orientations in \autoref{fig8}.

\begin{figure*}[t!]
\centering
\includegraphics[width=185 mm]{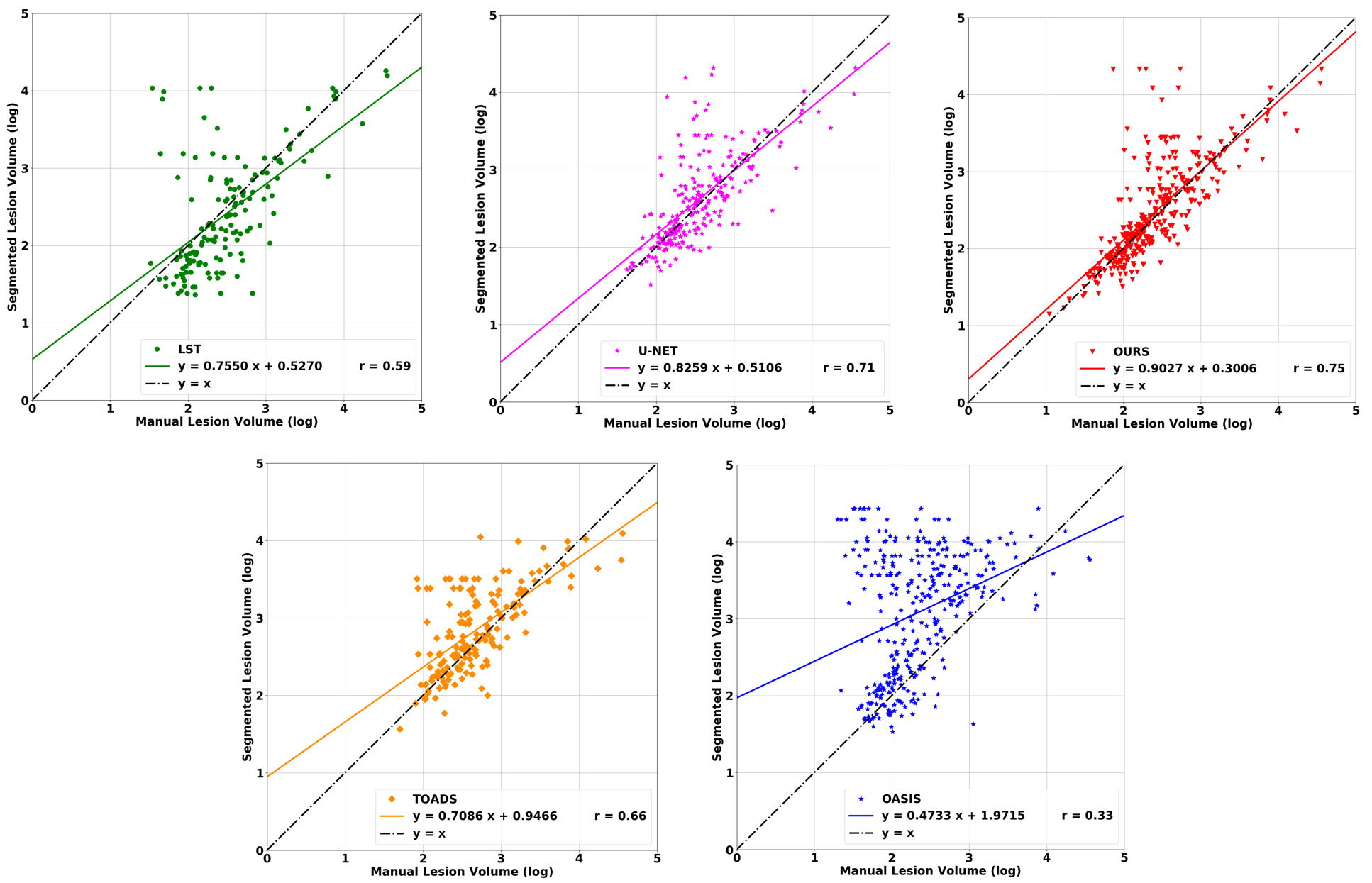}
\caption{Comparison of the lesion volumes produced by manual and automatic segmentation on the NRU dataset with different methods. Each point is associated with a single lesion. Colored (solid) lines indicate the correlation between manual and segmented lesion volumes. Black (dotted) lines indicate the ideal regression line. Slope, intercept, and Pearson's linear correlation (all with $p<<0.001$) between manual and estimated masks can also be seen for different methods.} 
\label{fig9}
\end{figure*}

\autoref{fig9} depicts the relationship between the volumes of all ground truth lesions and the corresponding estimated size for each evaluated method (one datapoint per lesion). With a qualitative evaluation, it can be seen that TOADS and OASIS methods tend to overestimate lesion volumes as many lesions are above the dashed black line, i.e., many lesions are estimated larger than they really are. On the contrary, LST method tends to underestimate the lesion sizes. U-Net and our method, on the contrary, produced lesions with size more comparable to the ground truth. However, with a quantitative analysis, our model produced the slope closest to unity (0.9027) together with the highest Pearson correlation coefficient (0.75), meaning our model provided the stronger global agreement between estimated and ground truth lesion volumes (note that a better agreement between lesion volumes does not mean the segmented and ground truth lesions better overlap -- the amount of overlap was measured with the \textit{DSC}).

\autoref{table4} shows the performance of the proposed model with respect to different combinations of modalities in the second experiment. The SB version of the proposed model used with one modality had noticeably better performance in almost all measures when using FLAIR modality. However, all modalities carry relevant information as better performance in most metrics was obtained when using a combination of modalities. In MB versions of the model, all possible two-branch and three-branch versions were considered. As shown in \autoref{table4}, two-branch versions including FLAIR modality showed a general better performance than the single-branch version using single modality. This emphasizes the importance of using FLAIR modality together with others (T1w and T2w). However, overall, a combination of all modalities in the three-branch version of the model showed the best general performance compared to the other versions of the network.

\section{Discussion and Conclusions}
In this work, we have designed an automated pipeline for MS lesion segmentation from multi-modal MRI data. The proposed model is a deep end-to-end 2D CNN consisting of a multi-branch downsampling network, MSFF blocks fusing the features from different modalities at different stages of the network, and MSFU blocks combining and upsampling multi-scale features.

When having insufficient training data in deep learning based approaches, which is very common in the medical domain, transfer learning has demonstrated to be an adequate solution \citep{chen2015automatic, chen2016dcan, hoo2016deep}. Not only it helps boosting the performance of the network but also it significantly reduces overfitting. Therefore, we used the parallel ResNet50s pre-trained on ImageNet as a multi-branch downsampling network while the other layers in MMFF and MSFU blocks were randomly initialized from a Gaussian distribution. We then fine-tuned the whole network on the given MS lesion segmentation task.

In brain image segmentation, a combination of MRI modalities overcomes the limitations of single modality approaches, allowing the models to provide more accurate segmentations \citep{kleesiek2016deep, moeskops2016automatic,Aslani_2019}. Unlike previously proposed deep networks \citep{brosch2016deep,Aslani_2019}, which stacked all modalities together as a single input, we designed a network with several downsampling branches, one branch for each individual modality. We believe that stacking all modalities together as a single input to a network is not an optimal solution since during the downsampling procedure, the details specific to the the most informative modalities can vanish when mixed with less informative modalities. On the contrary, the multi-branch approach allows the network to abstract higher-level features at different granularities specific to each modality. Independently of the ground truth used for training and testing the model, results in \autoref{table1} confirm our claim showing that a network with separate branches generated more accurate segmentations (e.g., \textit{DSC}=0.7649) than single-branch networks with all modalities stacked, as proposed by \citet{brosch2016deep} (e.g., \textit{DSC}=0.6844) and \citet{Aslani_2019} (e.g., \textit{DSC}=0.6980). Indeed, the mentioned methods (single-branch) generally obtained higher \textit{LTPR} values (e.g., 0.7455 and 0.7460) than multi-branch (e.g., 0.6697). However, they also obtained very high \textit{LFPR} values showing a significant overestimation of lesion volumes. The proposed method, instead, showed the best trade-off between \textit{LTPR} and \textit{LFPR}.

When examining the influence of different modalities, results in \autoref{table4} demonstratesin Table 4 demonstrated that the most important modality for that the most important modality for MS lesion segmentation was FLAIR (\textit{DSC}$>$0.65). This is likely due to the fact that FLAIR sequences benefit from CSF signal suppression and hence provide a higher image contrast between MS lesions and the surrounding normal appearing WM. Using all modalities together in a SB network (by concatenating them as single multi-channel input) and in a MB network (each modality as single input to each branch) showed good segmentation performance. This could be due to the combination of modalities helping the algorithm identifying additional information regarding the location of lesions. However, supporting our claim that stacking all modalities together as a single input to the network is not an optimal solution, top performance, indeed, was obtained in most measures with the MB network when using all available modalities, as can be seen in \autoref{table4}.

In deep CNNs, attributes from different layers include different information. Coarse layers are related to high-level semantic information (category specific), and shallow layers are related to low-level spatial information (appearance specific) \citep{long2015fully}, while middle layer attributes have shown a significant impact on segmentation performance \citep{ronneberger2015u}. Combining these multi-level attributes from the different stages of the network makes the representation richer than using single-level attributes, like in the CNN based method proposed by \citet{brosch2016deep}, where a single shortcut connection between the deepest and the shallowest layers was used. Our model, instead, includes several shortcut connections between all layers of the network, in order to combine multi-scale features from different stages of the network as inspired by U-Net architecture \citep{ronneberger2015u}. The results shown in \autoref{table1} suggest that the combination of multi-level features during the upsampling procedure helps the network exploiting more contextual information associated to the lesions. This could explain why the performance of our proposed model (\textit{DSC}=0.7649) is higher than the method proposed by \citet{brosch2016deep} (\textit{DSC}=0.6844).

Patch-based CNNs suffer from lacking spatial information about the lesions because of the patch size limitation. To deal with this problem, we proposed a whole-brain slice-based approach. Compared with patch-based methods \citep{valverde2017improving, ghafoorian2017deep}, we have shown that our model has better performance for most measures, as seen in \autoref{table2}. Although the CNN proposed by \citet{valverde2017improving} had the highest \textit{DSC} value among all, our method showed better performance regarding the \textit{LTPR} and \textit{LFPR}, which indicates that our model is robust in identifying the correct location of lesions. The method proposed by \citet{birenbaum2016longitudinal} has been optimized to have the highest \textit{LTPR}. However, their method showed significantly lower performance in \textit{LFPR}. Compared with this method, our method has better trade-off between \textit{LTPR} and \textit{LFPR}.

As mentioned in \citep{carass2017longitudinal}, manual delineation of MS lesions from MRI modalities is prone to intra- and inter-observer variability, which explains the relatively low \textit{DSC} between two experts delineating the same lesions ($\sim$0.73 for ISBI data as shown in \autoref{table1}). Automated methods are therefore expected to have a maximum performance in the same order of magnitude when comparing their generated segmentation with the rater's one. Accordingly, it is important to notice that, our model obtained a performance (\textit{DSC}) close to the experts agreement, as can be seen in \autoref{table1}.

The proposed method also has some limitations. We observed that the proposed pipeline is slightly slow in segmenting a 3D image since segmenting whole-brain slices takes a longer time compared to other CNN-based approaches \citep{roy2018multiple}. The time required to segment a 3D image is proportional to the size of the image and is based on the computational cost of three sequential steps: input features preparation \ref{input}, slice-level segmentation \ref{network architecture detail}, and 3D image reconstruction \ref{reconstruction}. In both the ISBI and NRU datasets, the average time for segmenting an input image with our model, including all 3 steps, was approximately 90 seconds.

A still open problem in MS lesion segmentation task is the identification of cortical and subcortical lesions. To this aim, we plan to use other MRI modalities such as double inversion recovery (DIR) sequences for the identification of cortical lesions, which benefits of the signal suppression from both CSF and WM. Moreover, we believe that introducing information from the tissue class could help improve the network identifying cortical, subcortical and white matter lesions. Therefore, we think that would be very promising to design a multi-task network for segmenting different parts of brain including different tissue types (WM, GM, CSF) and different types of MS lesions (including cortical lesions).

\begin{table*}[t]
\renewcommand{\thetable}{B.1}
\scriptsize
\centering
\caption{This table shows the results of the first experiment on the NRU dataset using our model as described in Section \ref{Experiments on the ISBI dataset)}. We implemented the same experiment using different methods for fusing output volumes (when merging the outputs from each plane orientation, and also when merging the outputs of models from different cross-validation folds). Mean values of \textit{DSC}, \textit{PPV}, \textit{LTPR}, \textit{LFPR}, \textit{VD}, \textit{SD} and \textit{HD} were measured for each method. Values in bold and italic indicate the first-best and second-best results.}
\begin{tabular}{lccccccc}
\hline
\multicolumn{1}{c}{Method} &DSC &PPV &LTPR &LFPR &VD &SD &HD \\ \hline
Majority Voting &\textbf{0.6655} &\textit{0.8032} &\textbf{0.4465} &\textit{0.0842} &\textbf{0.3372} &2.575 &\textbf{6.728}\\
Averaging &0.5883 &\textbf{0.8391} &0.3220 &\textbf{0.0788} &0.4625 &3.216 &\textit{8.503} \\
STAPLE \citep{warfield2004simultaneous} &\textit{0.6632} &0.7184 &\textit{0.3989} &0.0802 &\textit{0.3883} &\textbf{2.330} &8.629\\
\hline
\end{tabular}
\label{table8}
\end{table*}

\begin{table*}[t]
\renewcommand{\thetable}{C.1}
\scriptsize
\centering
\caption{This table indicates the performance of our trained model in the NRU dataset first experiment when using different ground truth masks as testing. Mean values of \textit{DSC}, \textit{PPV}, \textit{LTPR}, \textit{LFPR}, \textit{VD}, \textit{SD} and \textit{HD} were measured for each method. Values in bold and italic indicate the first-best and second-best results.}
\begin{tabular}{lccccccc}
\hline
\multicolumn{1}{c}{Method} &DSC &PPV &LTPR &LFPR &VD &SD &HD \\ \hline
Rater1 &\textbf{0.6827} &0.8010 &\textbf{0.5039} &0.0977 &0.3727 &\textbf{2.085} &\textbf{6.704}\\
Rater2 &0.6607 &0.7784 &0.4458 &0.0860 &0.3638 &2.511 &7.009 \\
Gold Standard (Consensus Mask) &0.6655 &\textbf{0.8032} &0.4465 &\textbf{0.0842} &\textbf{0.3372} &2.575 &6.728\\
\hline
\end{tabular}
\label{table9}
\end{table*}

Since the assessment of the disease burden from MRI of MS patients requires the quantification of the volume of hyperintense lesions on T2-weighted images, the final goal of the method proposed was to obtain an automatic and robust MS lesion segmentation tool. 
This will be particularly useful to facilitate scaling advanced MS analysis based on myelin imaging \citep{dayan2017mri} or multi-modal characterization of white matter tracts \citep{dayan2016profilometry} to large datasets. The long term goal, more generally, is the translation of this automatic method into a clinical tool. However, to be fully ready for clinical applications, the method should be also validated on healthy subjects and in a longitudinal framework.
The test on healthy subjects needs to be done to evaluate the amount of false positives generated by any approach on healthy brain scans. The experiments in a longitudinal framework are useful to assess the model reliability and capability to identify new, enlarged and stable lesions.
Moreover, still exploiting ISBI dataset, which includes longitudinal data, we could focus on leveraging this information to boost the performance of segmentation.

\begin{table}[b!]
\renewcommand{\thetable}{A.1}
\scriptsize
\centering
\caption{This table shows the implementation of first experiment in Section \ref{Experiments on the ISBI dataset)}. In this experiment, we evaluated our model using the ISBI dataset with available ground truth (training set with 5 subjects). We implemented a nested leave-one-subject-out cross-validation (3 subjects for training, 1 subject for validation, and 1 subject for testing). The numbers indicate the subject identifier.}
\begin{tabular}{ccc}
\hline
\multicolumn{1}{c}{Training} &Validation &Testing \\ \hline
1,2,3 & 4 & 5 \\
1,2,4 & 3 & 5 \\
1,3,4 & 2 & 5 \\
2,3,4 & 1 & 5 \\ \hline
1,2,3 & 5 & 4 \\
1,2,5 & 3 & 4 \\
1,3,5 & 2 & 4 \\
2,3,5 & 1 & 4 \\ \hline
1,2,4 & 5 & 3 \\
1,2,5 & 4 & 3 \\
1,4,5 & 2 & 3 \\
2,4,5 & 1 & 3 \\ \hline
1,3,4 & 5 & 2 \\
1,3,5 & 4 & 2 \\
1,4,5 & 3 & 2 \\
3,4,5 & 1 & 2 \\ \hline
2,3,4 & 5 & 1 \\
2,3,5 & 4 & 1 \\
2,4,5 & 3 & 1 \\
3,4,5 & 2 & 1 \\ \hline
\end{tabular}
\label{table5}
\end{table} 

\begin{table}[h!]
\renewcommand{\thetable}{A.2}
\scriptsize
\centering
\caption{This table shows the implementation of the second experiment in Section \ref{Experiments on the ISBI dataset)}. In this experiment, our model was evaluated using official ISBI test set including 14 subjects without publicly available ground truth. We trained our model doing a leave-one-subject-out cross-validation on whole training set (4 subject for training, 1 subject for validation, and 14 subject for testing). The numbers indicate the subject identifier.}
\label{my-label}
\begin{tabular}{ccc}
\hline
\multicolumn{1}{c}{Training} &Validation &Testing \\ \hline
1,2,3,4 & 5 & ISBI test set \\
1,2,3,5 & 4 & ISBI test set \\
1,2,4,5 & 3 & ISBI test set \\
1,3,4,5 & 2 & ISBI test set \\
2,3,4,5 & 1 & ISBI test set \\ \hline
\end{tabular}
\label{table6}
\end{table} 

\begin{table}[h!]
\renewcommand{\thetable}{A.3}
\scriptsize
\centering
\caption{This table gives detailed information regarding the training procedure for the first experiment in Section \ref{Experiment on the NRU dataset}. In this experiment, we implemented a nested 4-fold cross-validation over the whole NRU dataset including 37 subjects. [A-B @ C-D] denotes subjects A to B and C to D.}
\begin{tabular}{ccc}
\hline
\multicolumn{1}{c}{Training} &Validation &Testing \\ \hline
{[}17-37{]} & {[}10-16{]} & {[}1-9{]} \\
{[}10-16 @ 24-37{]} & {[}17-23{]} & {[}1-9{]} \\
{[}10-23 @ 31-37{]} & {[}24-30{]} & {[}1-9{]} \\
{[}10-30 @ 31-37{]} & {[}31-37{]} & {[}1-9{]} \\ \hline
{[}8-9 @ 19-37{]} & {[}1-7{]} & {[}10-18{]} \\
{[}1-7 @ 24-37{]} & {[}8-9 @ 19-23{]} & {[}10-18{]} \\
{[}1-9 @ 19-23 @ 31-37{]} & {[}24-30{]} & {[}10-18{]} \\
{[}1-9 @ 19-30{]} & {[}31-37{]} & {[}10-18{]} \\ \hline
{[}8-18 @ 28-37{]} & {[}1-7{]} & {[}19-27{]} \\
{[}1-7 @ 15-18 @ 27-37{]} & {[}8-14{]} & {[}19-27{]} \\
{[}1-14 @ 31-37{]} & {[}15-18 @ 28-30{]} & {[}19-27{]} \\
{[}1-18 @ 28-30{]} & {[}31-37{]} & {[}19-27{]} \\ \hline
{[}8-37{]} & {[}1-7{]} & {[}28-37{]} \\
{[}1-7 @ 15-27{]} & {[}8-14{]} & {[}28-37{]} \\
{[}1-14 @ 22-27{]} & {[}15-21{]} & {[}28-37{]} \\
{[}1-21{]} & {[}22-27{]} & {[}28-37{]} \\ \hline
\end{tabular}
\label{table7}
\end{table} 

\section*{Conflicts of Interest}
The authors have no conflicts of interest to declare.

\section*{Acknowledgments}
We  acknowledge the support of NVIDIA Corporation with the donation of the Titan Xp GPU used for this research.

\appendix

\section{Evaluation Protocols}
\label{app:A}
This appendix includes 3 tables that describe the training procedures in details related to Sections \ref{Experiments on the ISBI dataset)} and \ref{Experiment on the NRU dataset}.

\autoref{table5} and \autoref{table6} give detailed information about how we implemented training procedure on the ISBI dataset for the first and second experiments. \autoref{table7} describes the first and second experiments.  Table A.3 describes the nested 4-fold cross-validation training procedure applied on the NRU dataset in the first experiment. 

\section{Labels Aggregation}
\label{app:B}
In order to aggregate the outcomes of ensembles of labeling, beyond majority voting, we tested alternative well known label fusion methods. Specifically, we repeated the first experiment on NRU dataset as described in Section \ref{Experiments on the ISBI dataset)} substituting the majority vote framework with averaging and STAPLE (Simultaneous Truth and Performance Level) \citep{warfield2004simultaneous} methods, used to aggregate both the output volumes of the three plane orientations and the output volumes of the different models during cross-validation. \autoref{table8} indicates the performance of each method. Overall, majority voting had better performance than other methods. Therefore, we selected this method for all experiments.

\section{Rater Evaluation on NRU Dataset}
\label{app:C}
In the first NRU dataset experiment, beyond verifying the quality of the proposed model on the ground truth generated from the consensus of two experts, we also compared the performance with the ground truth from each individual experts. The rationale behind the experiment was to assess the consistency of the system across raters. \autoref{table9} shows the corresponding results. As expected from the high consensus between the masks provided by the two raters (as mentioned in Section~\ref{subsec:NRUdata}), our trained model using the gold standard mask (derived from the two raters' masks) showed comparable results when evaluated with either raters' masks or the consensus mask as ground truth.

\section*{}
\bibliographystyle{elsarticle-harv}
\bibliography{mybibfile.bib}

\begin{thebibliography}{57}
\expandafter\ifx\csname natexlab\endcsname\relax\def\natexlab#1{#1}\fi
\expandafter\ifx\csname url\endcsname\relax
  \def\url#1{\texttt{#1}}\fi
\expandafter\ifx\csname urlprefix\endcsname\relax\def\urlprefix{URL }\fi

\bibitem[{Abadi et~al.(2015)Abadi, Agarwal, Barham, Brevdo, Chen, Citro,
  Corrado, Davis, Dean, Devin, Ghemawat, Goodfellow, Harp, Irving, Isard, Jia,
  Jozefowicz, Kaiser, Kudlur, Levenberg, Man\'{e}, Monga, Moore, Murray, Olah,
  Schuster, Shlens, Steiner, Sutskever, Talwar, Tucker, Vanhoucke, Vasudevan,
  Vi\'{e}gas, Vinyals, Warden, Wattenberg, Wicke, Yu, and
  Zheng}]{tensorflow2015-whitepaper}
Abadi, M., Agarwal, A., Barham, P., Brevdo, E., Chen, Z., Citro, C., Corrado,
  G.~S., Davis, A., Dean, J., Devin, M., Ghemawat, S., Goodfellow, I., Harp,
  A., Irving, G., Isard, M., Jia, Y., Jozefowicz, R., Kaiser, L., Kudlur, M.,
  Levenberg, J., Man\'{e}, D., Monga, R., Moore, S., Murray, D., Olah, C.,
  Schuster, M., Shlens, J., Steiner, B., Sutskever, I., Talwar, K., Tucker, P.,
  Vanhoucke, V., Vasudevan, V., Vi\'{e}gas, F., Vinyals, O., Warden, P.,
  Wattenberg, M., Wicke, M., Yu, Y., Zheng, X., 2015. {TensorFlow}: Large-scale
  machine learning on heterogeneous systems. Software available from
  tensorflow.org.
\newline\urlprefix\url{https://www.tensorflow.org/}

\bibitem[{Andermatt et~al.(2017)Andermatt, Pezold, and
  Cattin}]{andermatt2017automated}
Andermatt, S., Pezold, S., Cattin, P.~C., 2017. Automated segmentation of
  multiple sclerosis lesions using multi-dimensional gated recurrent units. In:
  International MICCAI Brainlesion Workshop. Springer, pp. 31--42.

\bibitem[{Aslani et~al.(2019)Aslani, Dayan, Murino, and Sona}]{Aslani_2019}
Aslani, S., Dayan, M., Murino, V., Sona, D., 2019. Deep 2d encoder-decoder
  convolutional neural network for multiple sclerosis lesion segmentation in
  brain {MRI}. In: Brainlesion: Glioma, Multiple Sclerosis, Stroke and
  Traumatic Brain Injuries. Springer International Publishing, pp. 132--141.
\newline\urlprefix\url{https://doi.org/10.1007%2F978-3-030-11723-8_13}

\bibitem[{Birenbaum and Greenspan(2016)}]{birenbaum2016longitudinal}
Birenbaum, A., Greenspan, H., 2016. Longitudinal multiple sclerosis lesion
  segmentation using multi-view convolutional neural networks. In: Deep
  Learning and Data Labeling for Medical Applications. Springer, pp. 58--67.

\bibitem[{Brosch et~al.(2016)Brosch, Tang, Yoo, Li, Traboulsee, and
  Tam}]{brosch2016deep}
Brosch, T., Tang, L.~Y., Yoo, Y., Li, D.~K., Traboulsee, A., Tam, R., 2016.
  Deep 3d convolutional encoder networks with shortcuts for multiscale feature
  integration applied to multiple sclerosis lesion segmentation. IEEE
  transactions on medical imaging 35~(5), 1229--1239.

\bibitem[{Cabezas et~al.(2014)Cabezas, Oliver, Valverde, Beltran, Freixenet,
  Vilanova, Rami{\'o}-Torrent{\`a}, Rovira, and Llad{\'o}}]{cabezas2014boost}
Cabezas, M., Oliver, A., Valverde, S., Beltran, B., Freixenet, J., Vilanova,
  J.~C., Rami{\'o}-Torrent{\`a}, L., Rovira, {\`A}., Llad{\'o}, X., 2014.
  Boost: A supervised approach for multiple sclerosis lesion segmentation.
  Journal of neuroscience methods 237, 108--117.

\bibitem[{Carass et~al.(2017)Carass, Roy, Jog, Cuzzocreo, Magrath, Gherman,
  Button, Nguyen, Prados, Sudre, et~al.}]{carass2017longitudinal}
Carass, A., Roy, S., Jog, A., Cuzzocreo, J.~L., Magrath, E., Gherman, A.,
  Button, J., Nguyen, J., Prados, F., Sudre, C.~H., et~al., 2017. Longitudinal
  multiple sclerosis lesion segmentation: Resource and challenge. NeuroImage
  148, 77--102.

\bibitem[{Chen et~al.(2015)Chen, Dou, Ni, Cheng, Qin, Li, and
  Heng}]{chen2015automatic}
Chen, H., Dou, Q., Ni, D., Cheng, J.-Z., Qin, J., Li, S., Heng, P.-A., 2015.
  Automatic fetal ultrasound standard plane detection using knowledge
  transferred recurrent neural networks. In: International Conference on
  Medical Image Computing and Computer-Assisted Intervention. Springer, pp.
  507--514.

\bibitem[{Chen et~al.(2016)Chen, Qi, Yu, and Heng}]{chen2016dcan}
Chen, H., Qi, X., Yu, L., Heng, P.-A., 2016. Dcan: deep contour-aware networks
  for accurate gland segmentation. In: Proceedings of the IEEE conference on
  Computer Vision and Pattern Recognition. pp. 2487--2496.

\bibitem[{Chollet et~al.(2015)}]{chollet2015keras}
Chollet, F., et~al., 2015. Keras. \url{https://github.com/fchollet/keras}.

\bibitem[{Compston and Coles(2008)}]{compston2008multiple}
Compston, A., Coles, A., 2008. Multiple sclerosis. The Lancet 372~(9648),
  1502--1517.

\bibitem[{Dayan et~al.(2017)Dayan, Hurtado~R{\'u}a, Monohan, Fujimoto, Pandya,
  LoCastro, Vartanian, Nguyen, Raj, and Gauthier}]{dayan2017mri}
Dayan, M., Hurtado~R{\'u}a, S.~M., Monohan, E., Fujimoto, K., Pandya, S.,
  LoCastro, E.~M., Vartanian, T., Nguyen, T.~D., Raj, A., Gauthier, S.~A.,
  2017. Mri analysis of white matter myelin water content in multiple
  sclerosis: a novel approach applied to finding correlates of cortical
  thinning. Frontiers in neuroscience 11, 284.

\bibitem[{Dayan et~al.(2016)Dayan, Monohan, Pandya, Kuceyeski, Nguyen, Raj, and
  Gauthier}]{dayan2016profilometry}
Dayan, M., Monohan, E., Pandya, S., Kuceyeski, A., Nguyen, T.~D., Raj, A.,
  Gauthier, S.~A., 2016. Profilometry: a new statistical framework for the
  characterization of white matter pathways, with application to multiple
  sclerosis. Human brain mapping 37~(3), 989--1004.

\bibitem[{Deshpande et~al.(2015)Deshpande, Maurel, and
  Barillot}]{deshpande2015adaptive}
Deshpande, H., Maurel, P., Barillot, C., 2015. Adaptive dictionary learning for
  competitive classification of multiple sclerosis lesions. In: Biomedical
  Imaging (ISBI), 2015 IEEE 12th International Symposium on. IEEE, pp.
  136--139.

\bibitem[{Friedman et~al.(2000)Friedman, Hastie, Tibshirani,
  et~al.}]{friedman2000additive}
Friedman, J., Hastie, T., Tibshirani, R., et~al., 2000. Additive logistic
  regression: a statistical view of boosting (with discussion and a rejoinder
  by the authors). The annals of statistics 28~(2), 337--407.

\bibitem[{Ghafoorian et~al.(2017)Ghafoorian, Karssemeijer, Heskes, Bergkamp,
  Wissink, Obels, Keizer, de~Leeuw, van Ginneken, Marchiori,
  et~al.}]{ghafoorian2017deep}
Ghafoorian, M., Karssemeijer, N., Heskes, T., Bergkamp, M., Wissink, J., Obels,
  J., Keizer, K., de~Leeuw, F.-E., van Ginneken, B., Marchiori, E., et~al.,
  2017. Deep multi-scale location-aware 3d convolutional neural networks for
  automated detection of lacunes of presumed vascular origin. NeuroImage:
  Clinical 14, 391--399.

\bibitem[{Ghafoorian and Platel(2015)}]{ghafoorian2015convolutional}
Ghafoorian, M., Platel, B., 2015. Convolutional neural networks for ms lesion
  segmentation, method description of diag team. Proceedings of the 2015
  Longitudinal Multiple Sclerosis Lesion Segmentation Challenge, 1--2.

\bibitem[{Han et~al.(2016)Han, Lei, and Chen}]{han2016hep}
Han, X.-H., Lei, J., Chen, Y.-W., 2016. Hep-2 cell classification using
  k-support spatial pooling in deep cnns. In: International Workshop on
  Large-Scale Annotation of Biomedical Data and Expert Label Synthesis.
  Springer, pp. 3--11.

\bibitem[{Hashemi et~al.(2018)Hashemi, Salehi, Erdogmus, Prabhu, Warfield, and
  Gholipour}]{DBLP:journals/corr/abs-1803-11078}
Hashemi, S.~R., Salehi, S. S.~M., Erdogmus, D., Prabhu, S.~P., Warfield, S.~K.,
  Gholipour, A., 2018. Tversky as a loss function for highly unbalanced image
  segmentation using 3d fully convolutional deep networks. CoRR abs/1803.11078.
\newline\urlprefix\url{http://arxiv.org/abs/1803.11078}

\bibitem[{Havaei et~al.(2017)Havaei, Davy, Warde-Farley, Biard, Courville,
  Bengio, Pal, Jodoin, and Larochelle}]{havaei2017brain}
Havaei, M., Davy, A., Warde-Farley, D., Biard, A., Courville, A., Bengio, Y.,
  Pal, C., Jodoin, P.-M., Larochelle, H., 2017. Brain tumor segmentation with
  deep neural networks. Medical image analysis 35, 18--31.

\bibitem[{He et~al.(2016)He, Zhang, Ren, and Sun}]{he2016deep}
He, K., Zhang, X., Ren, S., Sun, J., 2016. Deep residual learning for image
  recognition. In: Proceedings of the IEEE conference on computer vision and
  pattern recognition. pp. 770--778.

\bibitem[{Hoo-Chang et~al.(2016)Hoo-Chang, Roth, Gao, Lu, Xu, Nogues, Yao,
  Mollura, and Summers}]{hoo2016deep}
Hoo-Chang, S., Roth, H.~R., Gao, M., Lu, L., Xu, Z., Nogues, I., Yao, J.,
  Mollura, D., Summers, R.~M., 2016. Deep convolutional neural networks for
  computer-aided detection: Cnn architectures, dataset characteristics and
  transfer learning. IEEE transactions on medical imaging 35~(5), 1285.

\bibitem[{Ioffe and Szegedy(2015)}]{ioffe2015batch}
Ioffe, S., Szegedy, C., 2015. Batch normalization: Accelerating deep network
  training by reducing internal covariate shift. arXiv preprint
  arXiv:1502.03167.

\bibitem[{Jain et~al.(2015)Jain, Sima, Ribbens, Cambron, Maertens, Van~Hecke,
  De~Mey, Barkhof, Steenwijk, Daams, et~al.}]{jain2015automatic}
Jain, S., Sima, D.~M., Ribbens, A., Cambron, M., Maertens, A., Van~Hecke, W.,
  De~Mey, J., Barkhof, F., Steenwijk, M.~D., Daams, M., et~al., 2015. Automatic
  segmentation and volumetry of multiple sclerosis brain lesions from mr
  images. NeuroImage: Clinical 8, 367--375.

\bibitem[{Jenkinson et~al.(2002)Jenkinson, Bannister, Brady, and
  Smith}]{jenkinson2002improved}
Jenkinson, M., Bannister, P., Brady, M., Smith, S., 2002. Improved optimization
  for the robust and accurate linear registration and motion correction of
  brain images. Neuroimage 17~(2), 825--841.

\bibitem[{Jenkinson and Smith(2001)}]{jenkinson2001global}
Jenkinson, M., Smith, S., 2001. A global optimisation method for robust affine
  registration of brain images. Medical image analysis 5~(2), 143--156.

\bibitem[{Kingma and Ba(2014)}]{DBLP:journals/corr/KingmaB14}
Kingma, D.~P., Ba, J., 2014. Adam: {A} method for stochastic optimization. CoRR
  abs/1412.6980.
\newline\urlprefix\url{http://arxiv.org/abs/1412.6980}

\bibitem[{Kleesiek et~al.(2016)Kleesiek, Urban, Hubert, Schwarz, Maier-Hein,
  Bendszus, and Biller}]{kleesiek2016deep}
Kleesiek, J., Urban, G., Hubert, A., Schwarz, D., Maier-Hein, K., Bendszus, M.,
  Biller, A., 2016. Deep mri brain extraction: a 3d convolutional neural
  network for skull stripping. NeuroImage 129, 460--469.

\bibitem[{LeCun et~al.(2015)LeCun, Bengio, and Hinton}]{lecun2015deep}
LeCun, Y., Bengio, Y., Hinton, G., 2015. Deep learning. nature 521~(7553), 436.

\bibitem[{LeCun et~al.(1998)LeCun, Bottou, Bengio, and
  Haffner}]{lecun1998gradient}
LeCun, Y., Bottou, L., Bengio, Y., Haffner, P., 1998. Gradient-based learning
  applied to document recognition. Proceedings of the IEEE 86~(11), 2278--2324.

\bibitem[{Li et~al.(2014)Li, Zhao, and Wang}]{li2014highly}
Li, H., Zhao, R., Wang, X., 2014. Highly efficient forward and backward
  propagation of convolutional neural networks for pixelwise classification.
  arXiv preprint arXiv:1412.4526.

\bibitem[{Liskowski and Krawiec(2016)}]{liskowski2016segmenting}
Liskowski, P., Krawiec, K., 2016. Segmenting retinal blood vessels with<? pub
  \_newline?> deep neural networks. IEEE transactions on medical imaging
  35~(11), 2369--2380.

\bibitem[{Liu et~al.(2017)Liu, Zhang, Song, Peng, and Cai}]{liu2017triple}
Liu, S., Zhang, D., Song, Y., Peng, H., Cai, W., 2017. Triple-crossing 2.5 d
  convolutional neural network for detecting neuronal arbours in 3d microscopic
  images. In: International Workshop on Machine Learning in Medical Imaging.
  Springer, pp. 185--193.

\bibitem[{Long et~al.(2015)Long, Shelhamer, and Darrell}]{long2015fully}
Long, J., Shelhamer, E., Darrell, T., 2015. Fully convolutional networks for
  semantic segmentation. In: Proceedings of the IEEE conference on computer
  vision and pattern recognition. pp. 3431--3440.

\bibitem[{Maier and Handels(2015)}]{maier2015ms}
Maier, O., Handels, H., 2015. Ms lesion segmentation in mri with random
  forests. Proc. 2015 Longitudinal Multiple Sclerosis Lesion Segmentation
  Challenge, 1--2.

\bibitem[{Milletari et~al.(2016)Milletari, Navab, and Ahmadi}]{milletari2016v}
Milletari, F., Navab, N., Ahmadi, S.-A., 2016. V-net: Fully convolutional
  neural networks for volumetric medical image segmentation. In: 3D Vision
  (3DV), 2016 Fourth International Conference on. IEEE, pp. 565--571.

\bibitem[{Moeskops et~al.(2016)Moeskops, Viergever, Mendrik, de~Vries, Benders,
  and I{\v{s}}gum}]{moeskops2016automatic}
Moeskops, P., Viergever, M.~A., Mendrik, A.~M., de~Vries, L.~S., Benders,
  M.~J., I{\v{s}}gum, I., 2016. Automatic segmentation of mr brain images with
  a convolutional neural network. IEEE transactions on medical imaging 35~(5),
  1252--1261.

\bibitem[{Nair and Hinton(2010)}]{nair2010rectified}
Nair, V., Hinton, G.~E., 2010. Rectified linear units improve restricted
  boltzmann machines. In: Proceedings of the 27th international conference on
  machine learning (ICML-10). pp. 807--814.

\bibitem[{Oishi et~al.(2008)Oishi, Zilles, Amunts, Faria, Jiang, Li, Akhter,
  Hua, Woods, Toga, et~al.}]{oishi2008human}
Oishi, K., Zilles, K., Amunts, K., Faria, A., Jiang, H., Li, X., Akhter, K.,
  Hua, K., Woods, R., Toga, A.~W., et~al., 2008. Human brain white matter
  atlas: identification and assignment of common anatomical structures in
  superficial white matter. Neuroimage 43~(3), 447--457.

\bibitem[{Rolak(2003)}]{rolak2003multiple}
Rolak, L.~A., 2003. Multiple sclerosis: it is not the disease you thought it
  was. Clinical Medicine and Research 1~(1), 57--60.

\bibitem[{Ronneberger et~al.(2015)Ronneberger, Fischer, and
  Brox}]{ronneberger2015u}
Ronneberger, O., Fischer, P., Brox, T., 2015. U-net: Convolutional networks for
  biomedical image segmentation. In: International Conference on Medical Image
  Computing and Computer-Assisted Intervention. Springer, pp. 234--241.

\bibitem[{Roth et~al.(2014)Roth, Lu, Seff, Cherry, Hoffman, Wang, Liu, Turkbey,
  and Summers}]{roth2014new}
Roth, H.~R., Lu, L., Seff, A., Cherry, K.~M., Hoffman, J., Wang, S., Liu, J.,
  Turkbey, E., Summers, R.~M., 2014. A new 2.5 d representation for lymph node
  detection using random sets of deep convolutional neural network
  observations. In: International conference on medical image computing and
  computer-assisted intervention. Springer, pp. 520--527.

\bibitem[{Roy et~al.(2018)Roy, Butman, Reich, Calabresi, and
  Pham}]{roy2018multiple}
Roy, S., Butman, J.~A., Reich, D.~S., Calabresi, P.~A., Pham, D.~L., 2018.
  Multiple sclerosis lesion segmentation from brain mri via fully convolutional
  neural networks. arXiv preprint arXiv:1803.09172.

\bibitem[{Schmidt et~al.(2012)Schmidt, Gaser, Arsic, Buck, F{\"o}rschler,
  Berthele, Hoshi, Ilg, Schmid, Zimmer, et~al.}]{schmidt2012automated}
Schmidt, P., Gaser, C., Arsic, M., Buck, D., F{\"o}rschler, A., Berthele, A.,
  Hoshi, M., Ilg, R., Schmid, V.~J., Zimmer, C., et~al., 2012. An automated
  tool for detection of flair-hyperintense white-matter lesions in multiple
  sclerosis. Neuroimage 59~(4), 3774--3783.

\bibitem[{Shiee et~al.(2010)Shiee, Bazin, Ozturk, Reich, Calabresi, and
  Pham}]{shiee2010topology}
Shiee, N., Bazin, P.-L., Ozturk, A., Reich, D.~S., Calabresi, P.~A., Pham,
  D.~L., 2010. A topology-preserving approach to the segmentation of brain
  images with multiple sclerosis lesions. NeuroImage 49~(2), 1524--1535.

\bibitem[{Simon et~al.(2006)Simon, Li, Traboulsee, Coyle, Arnold, Barkhof,
  Frank, Grossman, Paty, Radue, et~al.}]{simon2006standardized}
Simon, J., Li, D., Traboulsee, A., Coyle, P., Arnold, D., Barkhof, F., Frank,
  J., Grossman, R., Paty, D., Radue, E., et~al., 2006. Standardized mr imaging
  protocol for multiple sclerosis: Consortium of ms centers consensus
  guidelines. American Journal of Neuroradiology 27~(2), 455--461.

\bibitem[{Sled et~al.(1998)Sled, Zijdenbos, and Evans}]{sled1998nonparametric}
Sled, J.~G., Zijdenbos, A.~P., Evans, A.~C., 1998. A nonparametric method for
  automatic correction of intensity nonuniformity in mri data. IEEE
  transactions on medical imaging 17~(1), 87--97.

\bibitem[{Smith(2002)}]{smith2002fast}
Smith, S.~M., 2002. Fast robust automated brain extraction. Human brain mapping
  17~(3), 143--155.

\bibitem[{Steinman(1996)}]{steinman1996multiple}
Steinman, L., 1996. Multiple sclerosis: a coordinated immunological attack
  against myelin in the central nervous system. Cell 85~(3), 299--302.

\bibitem[{Sudre et~al.(2015)Sudre, Cardoso, Bouvy, Biessels, Barnes, and
  Ourselin}]{sudre2015bayesian}
Sudre, C.~H., Cardoso, M.~J., Bouvy, W.~H., Biessels, G.~J., Barnes, J.,
  Ourselin, S., 2015. Bayesian model selection for pathological neuroimaging
  data applied to white matter lesion segmentation. IEEE transactions on
  medical imaging 34~(10), 2079--2102.

\bibitem[{Sweeney et~al.(2013)Sweeney, Shinohara, Shiee, Mateen, Chudgar,
  Cuzzocreo, Calabresi, Pham, Reich, and Crainiceanu}]{sweeney2013oasis}
Sweeney, E.~M., Shinohara, R.~T., Shiee, N., Mateen, F.~J., Chudgar, A.~A.,
  Cuzzocreo, J.~L., Calabresi, P.~A., Pham, D.~L., Reich, D.~S., Crainiceanu,
  C.~M., 2013. Oasis is automated statistical inference for segmentation, with
  applications to multiple sclerosis lesion segmentation in mri. NeuroImage:
  clinical 2, 402--413.

\bibitem[{Tetteh et~al.(2018)Tetteh, Efremov, Forkert, Schneider, Kirschke,
  Weber, Zimmer, Piraud, and Menze}]{tetteh2018deepvesselnet}
Tetteh, G., Efremov, V., Forkert, N.~D., Schneider, M., Kirschke, J., Weber,
  B., Zimmer, C., Piraud, M., Menze, B.~H., 2018. Deepvesselnet: Vessel
  segmentation, centerline prediction, and bifurcation detection in 3-d
  angiographic volumes. arXiv preprint arXiv:1803.09340.

\bibitem[{Tomas-Fernandez and Warfield(2015)}]{tomas2015model}
Tomas-Fernandez, X., Warfield, S.~K., 2015. A model of population and subject
  (mops) intensities with application to multiple sclerosis lesion
  segmentation. IEEE transactions on medical imaging 34~(6), 1349--1361.

\bibitem[{Tseng et~al.(2017)Tseng, Lin, Hsu, and Huang}]{tseng2017joint}
Tseng, K.-L., Lin, Y.-L., Hsu, W., Huang, C.-Y., 2017. Joint sequence learning
  and cross-modality convolution for 3d biomedical segmentation. In: Computer
  Vision and Pattern Recognition (CVPR), 2017 IEEE Conference on. IEEE, pp.
  3739--3746.

\bibitem[{Vaidya et~al.(2015)Vaidya, Chunduru, Muthuganapathy, and
  Krishnamurthi}]{vaidya2015longitudinal}
Vaidya, S., Chunduru, A., Muthuganapathy, R., Krishnamurthi, G., 2015.
  Longitudinal multiple sclerosis lesion segmentation using 3d convolutional
  neural networks. Proceedings of the 2015 Longitudinal Multiple Sclerosis
  Lesion Segmentation Challenge, 1--2.

\bibitem[{Valverde et~al.(2017)Valverde, Cabezas, Roura,
  Gonz{\'a}lez-Vill{\`a}, Pareto, Vilanova, Rami{\'o}-Torrent{\`a}, Rovira,
  Oliver, and Llad{\'o}}]{valverde2017improving}
Valverde, S., Cabezas, M., Roura, E., Gonz{\'a}lez-Vill{\`a}, S., Pareto, D.,
  Vilanova, J.~C., Rami{\'o}-Torrent{\`a}, L., Rovira, {\`A}., Oliver, A.,
  Llad{\'o}, X., 2017. Improving automated multiple sclerosis lesion
  segmentation with a cascaded 3d convolutional neural network approach.
  NeuroImage 155, 159--168.

\bibitem[{Warfield et~al.(2004)Warfield, Zou, and
  Wells}]{warfield2004simultaneous}
Warfield, S.~K., Zou, K.~H., Wells, W.~M., 2004. Simultaneous truth and
  performance level estimation (staple): an algorithm for the validation of
  image segmentation. IEEE transactions on medical imaging 23~(7), 903--921.

\end{thebibliography}
\end{document}